\documentclass[pdflatex,sn-basic,iicol]{sn-jnl}

\usepackage{graphicx}
\usepackage{multirow}
\usepackage{amsmath,amssymb,amsfonts}
\usepackage{amsthm}
\usepackage{mathtools}
\usepackage{bm}
\usepackage{array}
\usepackage[table]{xcolor}
\usepackage{textcomp}
\usepackage{manyfoot}
\usepackage{booktabs}
\usepackage{algorithm}
\usepackage{algorithmic}
\usepackage{multicol}
\usepackage{pifont}
\usepackage{url}
\usepackage{verbatim}

\newcommand{\Tref}[1]{Table~\ref{#1}}
\newcommand{\Eref}[1]{Equation~(\ref{#1})}
\newcommand{\Fref}[1]{Fig.~\ref{#1}}
\newcommand{\Sref}[1]{Section~\ref{#1}}
\newcommand{\Aref}[1]{Algorithm~\ref{#1}}
\graphicspath{{figs/}}

\theoremstyle{thmstyleone}
\newtheorem{theorem}{Theorem}[section]
\newtheorem{proposition}[theorem]{Proposition}
\theoremstyle{thmstylethree}
\newtheorem{definition}[theorem]{Definition}
\theoremstyle{thmstyletwo}

\begin{document}

\title[Efficient Test-Time Adaptation]{Efficient Test-Time Adaptation through Latent Subspace Coefficients Search}

\author[1]{\fnm{Xinyu} \sur{Luo}}\email{x.luo@my.cityu.edu.hk}
\author[1]{\fnm{Jie} \sur{Liu}}\email{jliu.ee@my.cityu.edu.hk}
\author[1]{\fnm{Kecheng} \sur{Chen}}\email{ck.ee@my.cityu.edu.hk}
\author[2]{\fnm{Yihua} \sur{Shao}}\email{yihuajerry@gmail.com}
\author[1]{\fnm{Junyi} \sur{Yang}}\email{junyiyang8-c@my.cityu.edu.hk}
\author[1]{\fnm{Bo} \sur{Ding}}\email{bo.ding@my.cityu.edu.hk}
\author[3]{\fnm{Nicu} \sur{Sebe}}\email{sebe@disi.unitn.it}
\author[1]{\fnm{Arindam} \sur{Basu}}\email{arinbasu@cityu.edu.hk}
\author*[1]{\fnm{Haoliang} \sur{Li}}\email{haoliali@cityu.edu.hk}

\affil*[1]{\orgname{City University of Hong Kong}, \orgaddress{\city{Hong Kong SAR}, \country{China}}}
\affil[2]{\orgdiv{Institute of Automation}, \orgname{Chinese Academy of Sciences}, \orgaddress{\city{Beijing}, \country{China}}}
\affil[3]{\orgname{University of Trento}, \orgaddress{\city{Trento}, \country{Italy}}}

\abstract{
Real-world deployment often exposes vision models to distribution shifts, making test-time adaptation (TTA) important for reliable prediction. Existing TTA methods can be difficult to deploy on edge devices because they require backpropagation, activation storage, or test-time batches, which increase computation, memory, and state. We propose \textbf{ELaTTA} (\textit{Efficient Latent Test-Time Adaptation}), a forward-only, gradient-free framework for single-instance TTA. ELaTTA freezes the pretrained model and adapts each test input by searching a low-dimensional coefficient vector in a principal latent subspace computed offline from a source-domain cache. The basis is stored once and source data are discarded before deployment. At test time, CMA-ES performs covariance-adaptive Gaussian search, ranks candidate corrections by prediction entropy, and selects the best candidate within a fixed evaluation budget. On five visual benchmarks spanning corruption, natural, rendition, sketch, and cross-domain shifts, ELaTTA provides strong accuracy under strict and continual single-instance protocols. In the strict single-instance comparison, MEMO requires \emph{64.7$\times$} the computation and \emph{12.4$\times$} the peak memory of ELaTTA. A complementary keyword-spotting experiment shows that the same coefficient-search design transfers beyond vision. We further demonstrate ViT-Base adaptation on a Raspberry Pi 5, confirming the feasibility of deploying ELaTTA on an edge device.}


\maketitle

\section{Introduction}
The distribution encountered by a deployed vision model can differ substantially from the distribution used for training. Changes in image acquisition conditions, corruption processes, appearance, or scene context give rise to shifts between the source and test distributions, even when the task label space remains unchanged \citep{recht2019imagenet,hendrycks2019robustness,hendrycks2021many}. Such shifts can reduce the reliability of a fixed source model and motivate adaptation using unlabeled test inputs. Test-time adaptation (TTA) addresses this setting by using the test stream itself to improve predictions at inference time, without access to target labels \citep{sun2020test,darestani2022test,wang2025search,zhang2024unsupervised}. The amount of information available from the stream, and whether adaptation state is reset or carried forward, are therefore central aspects of an online TTA problem.

The need for adaptation is especially pronounced in on-device vision systems, where models operate close to the data source under strict latency, memory, and energy constraints \citep{li2024flexnn}. Local processing can reduce communication overhead and the need to transfer sensitive observations to a remote service, but it also requires adaptation to operate within limited device resources. Consequently, practical on-device TTA must balance adaptation effectiveness against the incremental computation, memory, and adaptation state introduced during inference.

These resource considerations are particularly relevant to \emph{gradient-based} TTA, a common approach that updates model parameters or auxiliary states during inference. Representative methods include entropy minimization with TENT \citep{wang2021tent}, sample-efficient adaptation with EATA \citep{niu2022efficient}, and consistency- or augmentation-based objectives such as MEMO and SPACE \citep{zhang2022memo,luo2025space}. These methods provide expressive adaptation mechanisms, but each gradient step requires backpropagation and intermediate activations, and repeated updates may introduce additional state and computation for every test input. Parameter- and memory-efficient variants reduce this cost in particular settings \citep{hong2023mecta,song2023ecotta,lee2024becotta,ma2025surgeon,tan2025uncertainty}. Nevertheless, adapting a general visual backbone under a tight per-sample budget remains challenging.

Gradient-free TTA avoids backpropagation and is consequently attractive for resource-constrained inference. However, existing gradient-free strategies may still rely on batch statistics or buffered test samples, as in normalization-based adaptation \citep{schneider2020improving,lim2023ttn}, or use model-specific classifier and logit adjustments \citep{iwasawa2021test,boudiaf2022parameter}. Such dependencies are difficult to reconcile with streams in which inputs arrive individually. Forward-only optimization also enables parameter adaptation. CAZO \citep{zhang2026curvature} uses curvature-aware zeroth-order gradient estimates to update lightweight adapters while freezing pretrained weights. Its parameter perturbations still require repeated network forward evaluations, making the cost of each adaptation step relevant even without backpropagation. Following the protocol distinctions in online TTA \citep{su2024revisiting,wang2025search}, we focus primarily on two single-instance settings. In \emph{strict single-instance} TTA, each test input is processed independently and the adaptation state is reset. In \emph{continual single-instance} TTA, inputs are processed sequentially and a limited adaptation state is carried across the stream. We additionally consider \emph{continual-batch} adaptation, in which multiple test inputs are processed together and the adaptation state is carried across batches. Together, these settings underscore the need for efficient TTA that preserves sufficient adaptation capacity under tight inference resource budgets, especially when inputs are processed individually.

\begin{figure}[!t]
\centering
\includegraphics[width=2.5in]{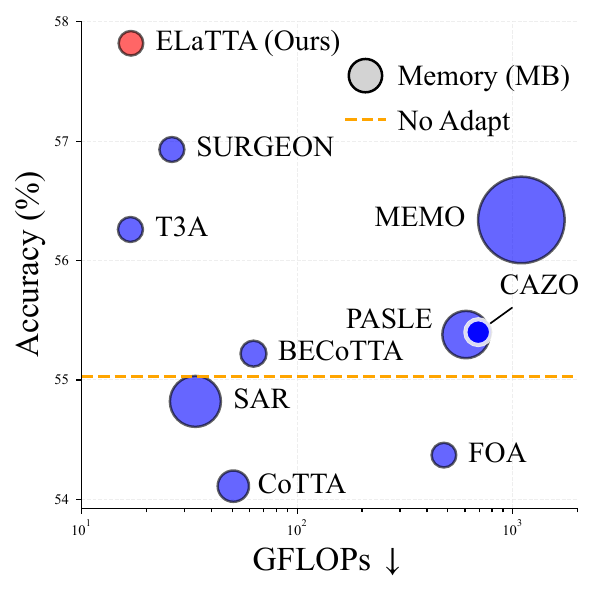}
\caption{Accuracy, computation, and peak memory of various TTA methods under the single-instance setting on ImageNet-C with ViT-Base model.}
\label{fig:comparison}
\end{figure}

To bridge this gap, we propose \underline{E}fficient \underline{La}tent \underline{T}est-\underline{T}ime \underline{A}daptation (\textbf{ELaTTA}), a gradient-free TTA method for edge deployment. ELaTTA combines three design choices: a source-derived principal subspace, a low-dimensional coefficient search, and covariance-adaptive Gaussian sampling. In an offline stage, we compute a truncated SVD from a small cache of source-training latents, retain the basis, and discard the source data. The basis is fixed during deployment and stored with negligible overhead (about \(0.015\%\) of the ViT-Base model size in our configuration). A small cache already gives an effective subspace, while larger caches make the result more consistent across cache draws.
At test time, ELaTTA keeps all network weights fixed and searches only a compact \(k\)-D coefficient vector. CMA-ES \citep{hansen2016cma} samples candidate corrections, ranks them by prediction entropy, and adapts the location, shape, and scale of the Gaussian search distribution within a fixed evaluation budget. The source basis determines the correction directions, and candidate feedback determines where the next evaluations are made.

Our main contributions are as follows:
\begin{enumerate}
\item{\textbf{Latent subspace coefficients TTA.} 
We introduce a latent coefficient adaptation paradigm that shifts TTA from updating high-dimensional weights or auxiliary parameters to searching a \emph{low-dimensional} latent subspace. 
This design confines each correction to source principal directions and keeps the backbone fixed across the test stream. The compact coefficient space supports per-input search with little persistent state, while the source-derived basis gives the search a task-relevant set of correction directions.}

\item{\textbf{Entropy minimization through adaptive distributional search.}
We use candidate entropy rankings to adapt a Gaussian search distribution within the latent subspace, then select the best sampled correction \emph{without auxiliary training losses or backpropagation}.
Our analysis characterizes the source-subspace geometry, the Gaussian-averaged entropy associated with coefficient sampling, and the loss and local prediction consistency of the selected candidate (see \Sref{sec:theory}). Component, source-cache, and calibration experiments examine how the search design translates into classification gains and reliable probabilities.}

\item{\textbf{Visual robustness and edge efficiency.}
We evaluate ELaTTA on five visual benchmarks and several backbone architectures under strict, continual single-instance, and continual-batch protocols, and use keyword spotting as a cross-modal supplementary task.
Across the visual experiments, ELaTTA is competitive or best under the stated protocol. In the strict single-instance comparison, MEMO requires \textbf{64.7$\times$} the computation and \textbf{12.4$\times$} the peak memory of ELaTTA. An end-to-end Raspberry Pi 5 experiment measures ELaTTA's practical latency and memory cost on an edge CPU. We further evaluate binary coefficient representations and fixed-point search to characterize the accuracy trade-offs of low-precision adaptation.}
\end{enumerate}

\section{Related Work}
\textbf{TTA Protocols.}
Test-time adaptation (TTA) improves a source model using unlabeled samples from a shifted test distribution, without requiring target annotations. Existing protocols differ in how test samples are presented and whether adaptation state is retained across samples \citep{liang2025comprehensive}. In strict single-instance TTA, each sample is processed independently and the adaptation state is reset after the prediction. In continual single-instance TTA, samples arrive sequentially and a limited state is carried from one sample to the next. In continual-batch TTA, samples are processed in batches and the adapted state is updated across batches. Although TTA covers broader settings, we focus on these protocols because they represent edge deployments with progressively larger budgets for persistent state and accumulated test information. These settings are often discussed under online TTA, where predictions must be produced as test samples or batches arrive. Batch size and state transfer therefore affect both adaptation quality and computational cost \citep{wang2025search}. Continual methods such as CoTTA \citep{wang2022continual}, SAR \citep{niutowards}, and BECoTTA \citep{lee2024becotta} are designed to update adaptation state over a test stream, whereas SITA \citep{khurana2021sita}, DUA \citep{mirza2022norm}, MEMO \citep{zhang2022memo}, and SPACE \citep{luo2025space} use normalization updates or augmented pseudo-batches for single-sample adaptation. These additional states or views can improve adaptation, but also increase computation and complicate deployment when inputs must be processed individually.

\noindent\textbf{Efficient TTA.}
Resource-constrained deployment requires adaptation methods to balance prediction improvement with per-sample computation, memory, and state. Gradient-based methods such as EATA-C \citep{tan2025uncertainty}, CoTTA \citep{wang2022continual}, SAR \citep{niutowards}, and BECoTTA \citep{lee2024becotta} reduce the number of updated parameters or control forgetting, but still require gradient computation, optimizer state, or repeated parameter updates. TinyTTA \citep{jia2024tinytta} reduces inference cost through an early-exit ensemble, but it changes the inference architecture and requires offline retraining. These requirements differ from our plug-and-play setting, in which the pretrained backbone and classifier remain fixed.

Gradient-free methods avoid backpropagation and are attractive for single-sample inference. Early approaches adapt normalization statistics from test data \citep{schneider2020improving,lim2023ttn}, while SITA \citep{khurana2021sita}, MixNorm \citep{hu2021mixnorm}, and NOTE \citep{gong2022note} use instance-level or augmented-view normalization updates. Other methods adapt classifier-side quantities, such as class prototypes in T3A \citep{iwasawa2021test} or logits in parameter-free correction \citep{boudiaf2022parameter}. Gradient-free methods differ in their adaptation variables and their requirements for test-time context. FOA \citep{niu2024test} searches input prompts using entropy and source-test feature-statistic alignment, requiring repeated encoder evaluations. CAZO \citep{zhang2026curvature} performs zeroth-order adapter updates with anisotropic perturbations shaped by a diagonal curvature proxy, estimated through an exponential moving average of squared gradient estimates. Its default objective combines prediction entropy with source-test feature-statistic alignment, and perturbing adapters within the encoder requires repeated forward evaluations through the subsequent layers. ELaTTA instead searches an additive correction in a source-derived latent subspace, using single-input entropy feedback and reusing the cached latent after one encoder pass.

\noindent\textbf{Latent Adaptation.}
Latent-space manipulation is widely used in image compression \citep{djelouah2019content,shen2023dec} and generative modeling \citep{shen2020interpreting,vahdat2021score}, but is less explored for TTA. A notable exception is Chen et~al.'s work \citep{chen2025test}, which refines latents for medical image segmentation using a latent CRF loss. While effective, it relies on backpropagation and is task-specific, limiting practicality for resource-constrained and general-purpose deployment. This gap motivates lightweight TTA methods that adapt by modifying latent representations efficiently, which our work seeks to address.

\section{Methodology}
\begin{figure*}[!t]
\centering
\includegraphics[width=\linewidth]{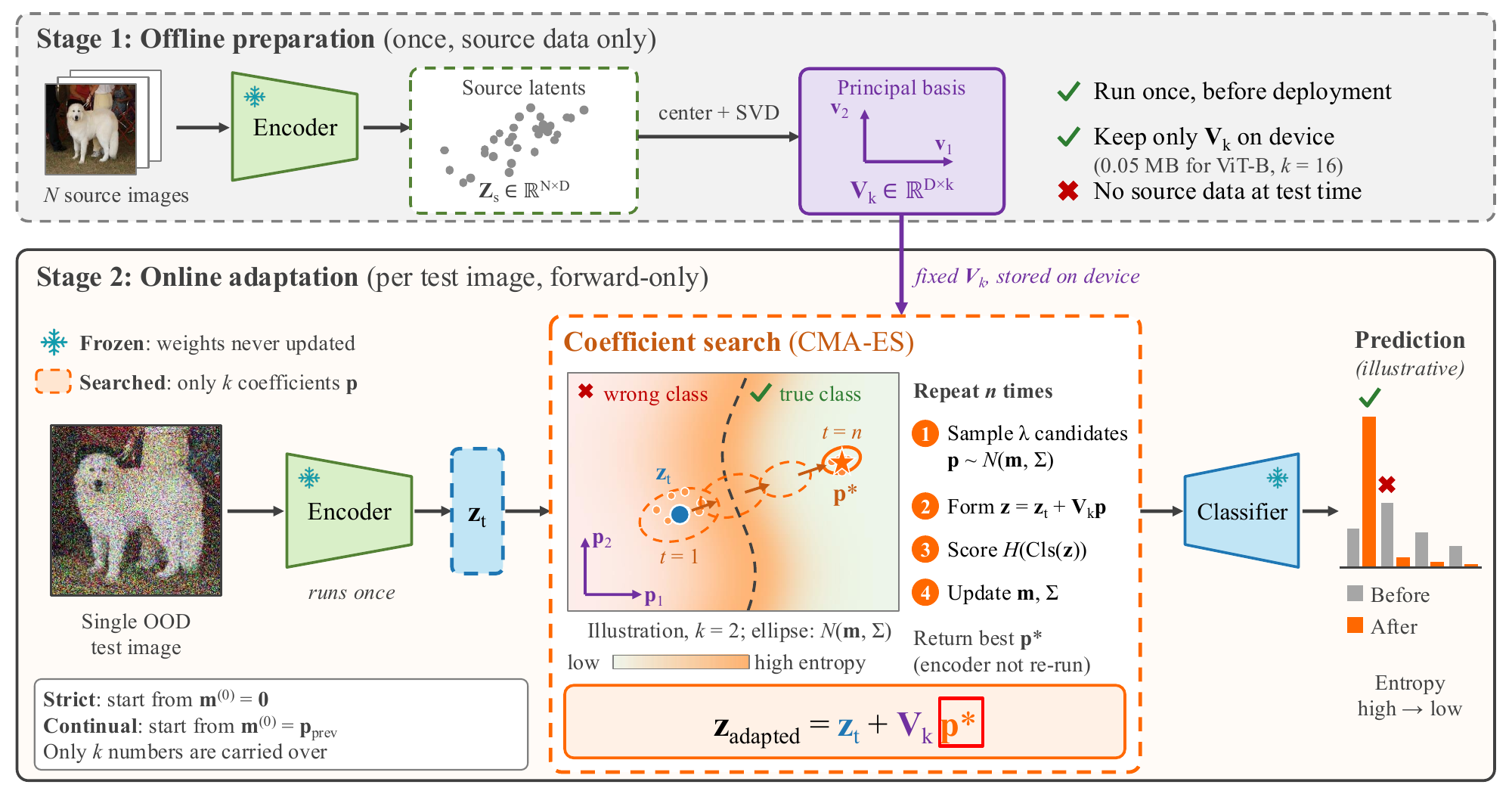}
\caption{\textbf{Overview of ELaTTA framework.} During offline preparation, we first compute a $k$-dimensional latent PC basis $\mathbf{V}_k$ from the source-domain latents. At test time, for a given single OOD sample, we perform a forward-only adaptation by searching for a low-dimensional coefficient
$\mathbf{p}\in\mathbb{R}^k$ within the source-induced subspace $\mathrm{span}(\mathbf{V}_k)$ and then selecting the update that minimizes prediction entropy. We use CMA-ES for this gradient-free optimization, enabling efficient test-time adaptation without costly backpropagation on edge devices.}
\label{fig:elatta}
\end{figure*}

\subsection{Problem Setting}
\label{sec:setup}
We consider unlabeled test images from a shifted distribution with the same class set as the source domain. Source images are used only offline to construct a latent subspace. At test time, adaptation uses the stored basis with the pretrained model weights fixed. We focus on \emph{strict single-instance} and \emph{continual single-instance} TTA, and also evaluate \emph{continual-batch} adaptation. Our goal is forward-only adaptation with low per-image computation and memory.

\noindent\textbf{Notation.}
We write \(f = \mathrm{Cls} \circ \mathrm{Enc}\), where \(\mathrm{Enc} : \mathcal{X} \to \mathbb{R}^{D}\) maps an image \(\mathbf{x}\) to its \(D\)-dimensional latent representation \(\mathbf{z} := \mathrm{Enc}(\mathbf{x})\). The classifier \(\mathrm{Cls}\), including softmax, outputs the probability vector \(\hat{\mathbf{y}} := \mathrm{Cls}(\mathbf{z})\) over \(C\) classes. We denote its pre-softmax logits by \(\mathbf{g}(\mathbf{z})\in\mathbb{R}^{C}\), so \(\mathrm{Cls}(\mathbf{z})=\mathrm{softmax}(\mathbf{g}(\mathbf{z}))\).

\subsection{ELaTTA: Latent Subspace Coefficients Adaptation}\label{sec:method_elatta}
\textbf{Overview.}
ELaTTA is a forward-only, gradient-free TTA framework for edge deployment.
It combines a source-derived principal subspace, low-dimensional latent coefficient search, and CMA-ES distributional optimization while keeping the network weights fixed.
The source basis restricts the correction directions, and CMA-ES uses entropy rankings to adapt the distribution of candidate coefficients. Each round uses the current input to guide the next round of sampling, and the best candidate determines the final prediction.

As illustrated in \Fref{fig:elatta}, we first extract a $k$-dimensional principal basis $\mathbf{V}_k\in\mathbb{R}^{D\times k}$ from a small cache of source latents.
At test time, given $\mathbf{z}_\text{t}=\mathrm{Enc}(\mathbf{x}_\text{t})$, we search a compact coefficient vector $\mathbf{p}\in\mathbb{R}^k$ with a label-free objective and adapt the latent as
\begin{equation}\label{eq:adaptation}
\mathbf{z}_{\text{adapted}} = \mathbf{z}_\text{t} + \mathbf{V}_k \mathbf{p}.
\end{equation}
We then predict $\hat{\mathbf{y}}=\mathrm{Cls}(\mathbf{z}_{\text{adapted}})$.

\noindent\textbf{Offline Preparation: Latent Subspace Construction.}
We build a compact latent subspace from source-domain features in a \emph{one-off offline stage, after which all source data is discarded and never accessed during deployment.}
Given a randomly selected set of $N$ samples from the source-domain training split, we extract their latents and stack them as a matrix $\mathbf{Z}_{\text{s}}\in\mathbb{R}^{N\times D}$, where each row is a latent representation vector.
We compute the source mean $\bm{\mu}_{\text{s}}$ and form the centered feature matrix $\widetilde{\mathbf{Z}}_{\text{s}}$; the mean is used only in this offline construction.
We then compute a rank-$k$ truncated SVD
\(\widetilde{\mathbf{Z}}_{\text{s}} \approx \mathbf{U}_k \bm{\Sigma}_k^{\!(z)} \mathbf{V}_k^\top,\)
where $\mathbf{V}_k\in\mathbb{R}^{D\times k}$ has orthonormal columns and spans the $k$-dimensional principal subspace capturing the dominant source variations (with $k\ll D$).
Once computed, we discard the raw latent matrix $\mathbf{Z}_{\text{s}}$ and retain only $\mathbf{V}_k$ for deployment. No source samples are accessed during TTA. As a low-dimensional basis, $\mathbf{V}_k$ does not retain individual source samples.
A small source cache is sufficient for strong performance at ImageNet scale in our experiments (see \Sref{sec:abl}).

\begin{algorithm}[t]
\caption{ELaTTA with $k$ Latent Subspace Coefficients}
\label{alg:tta}
\begin{algorithmic}

\STATE \textbf{Input:} Single test sample $\mathbf{x}$, encoder $\mathrm{Enc}$, classifier $\mathrm{Cls}$, latent PC basis $\mathbf{V}_k$, protocol, and iteration number $n$.
\STATE \textbf{Output:} Prediction $\hat{\mathbf{y}}^*$.

\STATE \textbf{Step 1: Generate latent representation.}
\STATE Obtain latent $\mathbf{z}_\text{t}$ by passing $\mathbf{x}$ through $\mathrm{Enc}$: $\mathbf{z}_\text{t} = \mathrm{Enc}(\mathbf{x})$.

\STATE \textbf{Step 2: Optimize latent adaptation.}
\STATE Initialize $\mathbf{m}^{(0)} \leftarrow \mathbf{0}$ for strict single-instance or $\mathbf{m}^{(0)} \leftarrow \mathbf{p}_{\text{prev}}$ for continual single-instance.
\FOR{$t = 1$ {\bfseries to} $n$}
    \STATE \textbf{Sampling:} Generate $\lambda$ candidate solutions.
    \STATE \textbf{Evaluation:} For each candidate $\mathbf{p}^{(t)}$, compute $\mathbf{z}_\text{adapted}$ by \Eref{eq:adaptation}. Obtain the output $\hat{\mathbf{y}} = \mathrm{Cls}(\mathbf{z}_\text{adapted})$ and compute the fitness via Shannon entropy.
    \STATE \textbf{Update:} Adapt CMA-ES internal parameters based on the top-performing candidates.
\ENDFOR

\STATE \textbf{Step 3: Return the final prediction.}
\STATE Return the $\mathbf{p}^*$ with the smallest fitness and its output $\hat{\mathbf{y}}^*$. In continual mode, set $\mathbf{p}_{\text{prev}} \leftarrow \mathbf{p}^*$.

\end{algorithmic}
\end{algorithm}

\noindent\textbf{Entropy-Based Test-Time Objective.}
At test time, ground-truth labels are unavailable. We therefore use Shannon entropy \citep{shannon1948mathematical} as a label-free confidence objective
\citep{grandvalet2004semi, wang2021tent, zhang2022memo, chen2025test} given as
$H(\hat{\mathbf{y}}) \triangleq -\sum_{c=1}^{C} \hat{y}_c \log \hat{y}_c$,
where \(\hat{y}_c\) is the probability of class $c$ and $C$ is the number of classes.

The adaptation does not update network parameters. We score each candidate coefficient $\mathbf{p}\in\mathbb{R}^k$ by its pointwise entropy,
\begin{equation}
\label{eq:tta_obj_pointwise}
\min_{\mathbf{p}\in\mathbb{R}^k}\;\; \mathcal{L}(\mathbf{p})
\quad \text{where}\quad
\mathcal{L}(\mathbf{p}) \triangleq
H\!\left(\mathrm{Cls}\!\left(\mathbf{z}_{\text{t}}+\mathbf{V}_k\mathbf{p}\right)\right).
\end{equation}
The update is restricted to \(\mathrm{span}(\mathbf{V}_k)\), which bounds the degrees of freedom of each adaptation.

\noindent\textbf{Searching with a Single-Input Entropy Signal.}
Near a decision boundary, small changes in the latent representation can change the predicted class. Entropy favors confident predictions, including confident errors under distribution shift \citep{lee2024entropy}. The search strategy therefore matters when adapting from a single unlabeled input. ELaTTA evaluates a population of corrections in the source subspace and uses their relative entropies to guide subsequent sampling. This provides feedback across several candidate locations while keeping the search within a compact space and a fixed evaluation budget.

\noindent\textbf{Gaussian-Averaged Entropy: A Distributional View.}
CMA-ES maintains a Gaussian distribution \(\mathbf{p}\sim\mathcal{N}(\mathbf{m},\mathbf{\Sigma})\) over candidate coefficients. Its average entropy is
\begin{equation}
\label{eq:tta_obj_smoothed}
J(\mathbf{m},\mathbf{\Sigma})
\triangleq \mathbb{E}_{\mathbf{p}\sim\mathcal{N}(\mathbf{m},\mathbf{\Sigma})}
\Big[
H\!\big(\mathrm{Cls}(\mathbf{z}_{\text{t}}+\mathbf{V}_k\mathbf{p})\big)
\Big].
\end{equation}
For fixed \(\mathbf{\Sigma}\), \(J\) is a Gaussian-smoothed version of \(\mathcal{L}\) as a function of \(\mathbf{m}\). It summarizes the entropy across the sampled neighborhood, whose location and extent are determined by \(\mathbf{m}\) and \(\mathbf{\Sigma}\). This distributional view connects coefficient sampling to the latent perturbations analyzed below. Candidate entropy ranks guide Gaussian updates.

\noindent\textbf{Gradient-Free Distributional Search via CMA-ES.}
CMA-ES updates the search distribution using only forward evaluations. At each iteration, it samples \(\lambda\) candidates from the current Gaussian, evaluates their pointwise entropies \(\mathcal{L}(\mathbf{p}_i)\), and updates the distribution from the ranked candidates. Across all evaluated candidates during the \(n\) iterations, we keep the best coefficient
$\mathbf{p}^* \triangleq \arg\min_{\mathbf{p}\in \mathcal{S}}\; \mathcal{L}(\mathbf{p})$,
where \(\mathcal{S}\) denotes the set of all sampled coefficients. We then form the adapted latent via \Eref{eq:adaptation} and output the corresponding prediction.
The final prediction uses this best candidate within the available evaluation budget. In continual-batch mode, the same search uses one coefficient shared by the batch, and the next batch is initialized from the previous batch solution. The overall algorithm is shown in \Aref{alg:tta}.

\subsection{Theoretical Analysis}
\label{sec:theory}
We analyze ELaTTA through three properties: the directions accessible to latent correction, the Gaussian average associated with coefficient sampling, and the loss and local prediction consistency of the selected candidate. Together, these properties describe how the source basis, search distribution, and output selection shape adaptation.

\noindent\textbf{Effect of Subspace-Constrained Latent Adaptation.}
Let \(\mathbf{z}_\mathrm{c}\in\mathbb{R}^D\) be a clean latent feature and \(\mathbf{z}_\text{t}=\mathbf{z}_\mathrm{c}+\bm{\xi}\) the corresponding test latent. For the orthonormal source basis \(\mathbf{V}_k\), define the orthogonal projector \(\mathbf{P}=\mathbf{V}_k\mathbf{V}_k^\top\) and decompose \(\bm{\xi}=\bm{\xi}_\parallel+\bm{\xi}_\perp\), where \(\bm{\xi}_\parallel=\mathbf{P}\bm{\xi}\) and \(\bm{\xi}_\perp=(\mathbf{I}-\mathbf{P})\bm{\xi}\) \citep{golub2013matrix, horn2012matrix}.

\begin{proposition}[Subspace-Constrained Error Decomposition]
\label{prop:subspace}
For \(\mathbf{z}_{\text{adapted}}=\mathbf{z}_\text{t}+\mathbf{V}_k\mathbf{p}\),
\begin{equation}
\label{eq:subspace_decomp}
\|\mathbf{z}_{\text{adapted}}-\mathbf{z}_\mathrm{c}\|^2
=\|\bm{\xi}_\parallel+\mathbf{V}_k\mathbf{p}\|^2
+\|\bm{\xi}_\perp\|^2.
\end{equation}
Thus, the orthogonal component \(\bm{\xi}_\perp\) is unaffected by the coefficient search.
\end{proposition}

\begin{proof}
Since \(\mathbf{P}\) is an orthogonal projector, both \(\bm{\xi}_\parallel\) and \(\mathbf{V}_k\mathbf{p}\) lie in \(\mathrm{span}(\mathbf{V}_k)\), while \(\bm{\xi}_\perp\) is orthogonal to this subspace. Therefore,
\[
\mathbf{z}_{\text{adapted}}-\mathbf{z}_\mathrm{c}
=(\bm{\xi}_\parallel+\mathbf{V}_k\mathbf{p})+\bm{\xi}_\perp
\]
is a sum of orthogonal terms, giving \Eref{eq:subspace_decomp}. The second term is independent of \(\mathbf{p}\).
\end{proof}

The basis fixes the directions accessible to coefficient search, and the coefficients determine the correction within those directions. This decomposition identifies both the capacity and the geometric limitation of latent correction. We examine the effect of the source basis on classification in \Sref{sec:abl}.

\noindent\textbf{Distribution Smoothing in Coefficient Space.}
Let \(\mathcal{L}(\mathbf{p})\) be the pointwise entropy in \Eref{eq:tta_obj_pointwise} and \(J(\mathbf{m},\mathbf{\Sigma})\) the Gaussian-averaged entropy in \Eref{eq:tta_obj_smoothed}.

\begin{proposition}[Score-function gradient and induced rank-\(k\) smoothing]
\label{prop:smoothing}
Assume that \(\mathbf{\Sigma}\) is positive definite and differentiation under the integral sign is valid. With covariance held fixed, the gradient of \(J\) with respect to the mean satisfies
\begin{equation}
\label{eq:grad_J_mean}
\nabla_{\mathbf{m}}J(\mathbf{m},\mathbf{\Sigma})
=\mathbb{E}_{\mathbf{p}\sim\mathcal{N}(\mathbf{m},\mathbf{\Sigma})}
\Big[\mathcal{L}(\mathbf{p})\,\mathbf{\Sigma}^{-1}(\mathbf{p}-\mathbf{m})\Big].
\end{equation}
Moreover, sampling \(\mathbf{p}\) induces
\(\mathbf{z}(\mathbf{p})=\mathbf{z}_\text{t}+\mathbf{V}_k\mathbf{p}\sim\mathcal{N}(\mathbf{z}_\text{t}+\mathbf{V}_k\mathbf{m},\,\mathbf{V}_k\mathbf{\Sigma}\mathbf{V}_k^\top)\), whose covariance is rank-\(k\).
\end{proposition}

\begin{proof}
Write \(q(\mathbf{p};\mathbf{m},\mathbf{\Sigma})\) for the Gaussian density. Then \(J=\int\mathcal{L}(\mathbf{p})q(\mathbf{p};\mathbf{m},\mathbf{\Sigma})\,d\mathbf{p}\), which is a Gaussian convolution of \(\mathcal{L}\) as a function of \(\mathbf{m}\) for fixed \(\mathbf{\Sigma}\). Differentiating the density gives
\[
\nabla_{\mathbf{m}}q
=q\,\mathbf{\Sigma}^{-1}(\mathbf{p}-\mathbf{m}).
\]
Differentiation under the integral sign therefore yields
\begin{align*}
\nabla_{\mathbf{m}}J
&=\int\mathcal{L}(\mathbf{p})\,\nabla_{\mathbf{m}}q\,d\mathbf{p}\\
&=\int\mathcal{L}(\mathbf{p})\,q\,\mathbf{\Sigma}^{-1}(\mathbf{p}-\mathbf{m})\,d\mathbf{p},
\end{align*}
which gives \Eref{eq:grad_J_mean}. The affine map \(\mathbf{z}(\mathbf{p})=\mathbf{z}_\text{t}+\mathbf{V}_k\mathbf{p}\) preserves Gaussianity, with mean \(\mathbf{z}_\text{t}+\mathbf{V}_k\mathbf{m}\) and covariance \(\mathbf{V}_k\mathbf{\Sigma}\mathbf{V}_k^\top\). Because \(\mathbf{V}_k\) has \(k\) orthonormal columns and \(\mathbf{\Sigma}\) is positive definite, this covariance has rank \(k\), and the latent perturbations are restricted to \(\mathrm{span}(\mathbf{V}_k)\).
\end{proof}

The induced covariance specifies the directions and scales explored around the adapted mean. CMA-ES changes this neighborhood using ranked candidate entropies. The gradient identity characterizes \(J\), while candidate rankings drive the implemented search updates; the lowest-entropy candidate supplies the final prediction.

\noindent\textbf{Candidate Selection and Local Prediction Consistency.}
\Aref{alg:tta} returns the sampled coefficient with the smallest pointwise entropy. The following result relates its expected entropy to the average entropy of the search distributions and gives a condition for agreement with the prediction at a search mean.

\begin{proposition}[Expected-loss bound and local margin transfer]
\label{prop:bridge}
Suppose each CMA-ES population is sampled from its current Gaussian search distribution, and let \(\mathbf{p}^*\) be the best sample across all iterations. Then
(i) the expected pointwise loss satisfies
\begin{equation}
\label{eq:expected_loss_bound}
\mathbb{E}[\mathcal{L}(\mathbf{p}^*)]
\le\min_t\mathbb{E}[J(\mathbf{m}_t,\mathbf{\Sigma}_t)].
\end{equation}
(ii) fix a search mean \(\mathbf{m}\) and let \(c_1,c_2\) be the top two classes at \(\mathbf{z}_{\mathbf{m}}=\mathbf{z}_\text{t}+\mathbf{V}_k\mathbf{m}\), with positive logit margin \(\gamma_{\mathbf{m}}=g_{c_1}(\mathbf{z}_{\mathbf{m}})-g_{c_2}(\mathbf{z}_{\mathbf{m}})>0\). Suppose every logit \(g_c\) is \(K\)-Lipschitz, with \(K>0\), on a neighborhood \(\mathcal U\) containing both \(\mathbf{z}_{\mathbf{m}}\) and \(\mathbf{z}_{\mathbf{p}^*}=\mathbf{z}_\text{t}+\mathbf{V}_k\mathbf{p}^*\). The prediction at \(\mathbf{m}\) is preserved at \(\mathbf{p}^*\) when
\begin{equation}
\label{eq:margin_transfer}
\|\mathbf{p}^*-\mathbf{m}\|_2<\frac{\gamma_{\mathbf{m}}}{2K}.
\end{equation}
\end{proposition}

\begin{proof}
\textit{Part (i).} For every iteration \(t\), the globally selected candidate satisfies the pointwise inequality
\[
\mathcal{L}(\mathbf{p}^*)
\le\min_i\mathcal{L}(\mathbf{p}_i^{(t)})
\le\frac{1}{\lambda}\sum_{i=1}^{\lambda}\mathcal{L}(\mathbf{p}_i^{(t)}).
\]
Conditioned on the current search parameters, each candidate has distribution \(\mathcal{N}(\mathbf{m}_t,\mathbf{\Sigma}_t)\). Thus,
\begin{align*}
\mathbb{E}\big[\mathcal{L}(\mathbf{p}^*)\mid\mathbf{m}_t,\mathbf{\Sigma}_t\big]
&\le\frac{1}{\lambda}\sum_{i=1}^{\lambda}
\mathbb{E}\big[\mathcal{L}(\mathbf{p}_i^{(t)})\mid\mathbf{m}_t,\mathbf{\Sigma}_t\big]\\*
&=J(\mathbf{m}_t,\mathbf{\Sigma}_t).
\end{align*}
Taking unconditional expectations by the tower property gives \(\mathbb{E}[\mathcal{L}(\mathbf{p}^*)]\le\mathbb{E}[J(\mathbf{m}_t,\mathbf{\Sigma}_t)]\) for every \(t\), yielding \Eref{eq:expected_loss_bound}. This argument allows the search parameters to depend on previous rounds.

\textit{Part (ii).} Let \(d=\|\mathbf{z}_{\mathbf{p}^*}-\mathbf{z}_{\mathbf{m}}\|_2\). The Lipschitz assumption implies
\begin{align*}
g_{c_1}(\mathbf{z}_{\mathbf{p}^*})&\ge g_{c_1}(\mathbf{z}_{\mathbf{m}})-Kd,\\
g_c(\mathbf{z}_{\mathbf{p}^*})&\le g_c(\mathbf{z}_{\mathbf{m}})+Kd.
\end{align*}
Since \(c_2\) is the strongest competitor at \(\mathbf{z}_{\mathbf{m}}\), for every \(c\ne c_1\),
\begin{align*}
g_{c_1}(\mathbf{z}_{\mathbf{p}^*})-g_c(\mathbf{z}_{\mathbf{p}^*})
&\ge g_{c_1}(\mathbf{z}_{\mathbf{m}})-g_c(\mathbf{z}_{\mathbf{m}})-2Kd\\
&\ge\gamma_{\mathbf{m}}-2Kd.
\end{align*}
Orthonormality of \(\mathbf{V}_k\) gives \(d=\|\mathbf{V}_k(\mathbf{p}^*-\mathbf{m})\|_2=\|\mathbf{p}^*-\mathbf{m}\|_2\). Under \Eref{eq:margin_transfer}, the lower bound is positive for all competing classes, so \(c_1\) remains the top-1 prediction.
\end{proof}

The first part quantifies the entropy advantage of selecting the best evaluated candidate relative to the population average. The second part gives a local agreement condition: a larger logit margin permits a greater displacement from the search mean. Since \(\mathbf{V}_k\) preserves coefficient distances, the condition is expressed directly in the low-dimensional search space. We evaluate the resulting classification effects in \Sref{sec:abl}.

\subsection{Complexity and Deployment}
\label{sec:complexity_deployment}
\textbf{Offline statistic and storage.}
The basis $\mathbf{V}_k\in\mathbb{R}^{D\times k}$ is computed once offline, after which the raw features are discarded. It remains fixed during deployment, and no source samples are accessed at test time.
For ViT-B with $D=768$ and $k=16$, storing $\mathbf{V}_k$ in FP32 requires
$
768\times16\times4\ \text{bytes}\approx 0.05\ \text{MB}.
$
Compared with the $\approx 330.28$ MB FP32 model, this is
$
\frac{0.05}{330.28}\times 100\% \approx 0.015\%,
$
of the model size. This quantifies the persistent storage overhead introduced by ELaTTA.

\noindent\textbf{Online overhead.}
With population size $\lambda$ and $n$ iterations, ELaTTA performs $E=\lambda n$ candidate evaluations per test sample. The encoder runs once to cache $\mathbf z_{\text{t}}$, and each candidate then uses the classifier with $\mathbf z_{\text{t}}+\mathbf{V}_k\mathbf{p}$. Thus, the network-forward component scales as $F_{\mathrm{Enc}}+E F_{\mathrm{Cls}}$. Latent updates and entropy evaluation add $\mathcal{O}(EDk)$ and $\mathcal{O}(EC)$ operations, respectively. A dense-covariance CMA-ES implementation adds $\mathcal{O}(Ek^2)$ distribution-update cost. The best candidate and its output are retained, so no extra classifier pass is needed. Candidate evaluations for one test image can run in parallel, which does not change the test-time batch protocol. We report empirical latency and peak memory in \Sref{sec:computation}.

\section{Experiments}
\subsection{Experimental Setup and Protocols}
\label{sec:exp_setup}

\textbf{Datasets.}
We evaluate visual classification on five benchmarks covering different forms of distribution shift. ImageNet-C \citep{hendrycks2019robustness} applies 15 corruption types, grouped into noise, blur, weather, and digital categories, to the 50,000 ImageNet-1k validation images. We evaluate all corruption types at severity 5. ImageNet-V2 \citep{recht2019imagenet} assesses natural distribution shift using newly collected images from the same 1,000 classes. ImageNet-R \citep{hendrycks2021many} contains approximately 30,000 artistic renditions from 200 ImageNet classes, while ImageNet-Sketch \citep{wang2019learning} contains approximately 50,000 sketch images spanning the 1,000 classes. These benchmarks assess adaptation to changes in appearance, texture, and visual representation. DomainNet-126 \citep{peng2019moment} contains 126 classes across painting, clipart, real, and sketch domains and extends the evaluation to cross-domain appearance and style shifts.

\noindent\textbf{Backbones.}
We use ViT-Base \citep{dosovitskiyimage} pretrained on ImageNet-1k \citep{ILSVRC15} for the main ImageNet comparisons, and ResNet-50 \citep{he2016deep}, EfficientNet-B0 \citep{tan2019efficientnet}, and MobileNet-V4 \citep{qin2024mobilenetv4} for additional architecture comparisons. DomainNet-126 uses ResNet-50. A keyword-spotting (KWS) experiment with an LSTM \citep{yang202533} evaluates applicability beyond vision.

\noindent\textbf{Evaluation protocols.}
In \emph{strict single-instance} TTA, each input is processed independently with batch size one, and adaptation state is reset after each sample. In \emph{continual single-instance} TTA, state is retained across successive inputs. In \emph{continual-batch} TTA, state is retained across batches of 64 inputs. Unless specified otherwise, experiments use the strict protocol. We report classification accuracy (\%) and express accuracy differences in percentage points.

\noindent\textbf{Baselines.}
The main visual comparisons include gradient-free methods T3A \citep{iwasawa2021test}, FOA \citep{niu2024test}, and CAZO \citep{zhang2026curvature}, and gradient-based methods CoTTA \citep{wang2022continual}, MEMO \citep{zhang2022memo}, SAR \citep{niutowards}, BECoTTA \citep{lee2024becotta}, PASLE \citep{huselective}, and SURGEON \citep{ma2025surgeon}. In strict evaluation, these methods use batch size one and reset their adaptation state after each sample. For SAR, the official one-step implementation updates the model using the current input but returns the output from the forward pass before that update. Under strict single-instance evaluation, this would exclude the current-input update from the reported prediction because the state is reset after the sample. We therefore retain the official SAR update rule but modify the evaluation wrapper to complete the update before collecting the reported prediction for the current input. BECoTTA is adapted to classification by applying its MoDE modules to the classification backbone. Other baselines use their official implementations and recommended hyperparameters. Continual and architecture-specific baselines are introduced in the corresponding subsections.

\begin{table*}[ht]
\caption{Five backbone models and their hidden size $D$ (first dimension of the latent PC basis $\mathbf{V}_k$).}
\label{tab:experiment_backbones}
\centering
\begin{tabular}{l c c c c c}
\toprule
Model &  ViT-Base & ResNet-50 & EfficientNet-B0 & MobileNet-V4 & LSTM\\ 
\midrule
$D$ &  768 & 2048 & 1280 & 1280 & 32 \\
\bottomrule
\end{tabular}
\end{table*}

\noindent\textbf{ELaTTA configuration.}
We compute $\mathbf{V}_k$ offline from source latents and keep it fixed at test time. For ImageNet, we use a fixed reference cache of 50,000 images randomly sampled from the ImageNet-1k training split. Visual experiments use $k=16$ and $n=8$ unless specified otherwise, and  KWS uses $k=2$ and $n=2$. The population size follows $\lambda=\lfloor 4+3\ln(k)\rfloor$. \Tref{tab:experiment_backbones} lists the latent dimension $D$. The model weights remain fixed throughout adaptation.

\subsection{Main Visual Results}
\label{sec:main_visual}
\textbf{Strict single-instance adaptation on ImageNet-C.}
As shown in \Tref{tab:ic_1}, ELaTTA achieves the best average accuracy of 57.82\% and delivers strong performance across most corruption types, indicating robust adaptation under diverse shifts.
Under this strict setting, methods that depend on batch/stream statistics or stabilized pseudo-targets tend to be less effective. FOA benefits from richer test-time context, CoTTA's EMA teacher can be noisy when updated from single-sample pseudo-labels, and SAR's reliable-sample selection and sharpness-aware updates are difficult to realize with batch-$1$ inputs.
History-based methods such as T3A are also disadvantaged because their support-set updates make predictions order-dependent; enforcing per-instance independence largely weakens the adaptation signal. CAZO reaches 55.40\%, only 0.37 points above No Adapt. Resetting its adapter and curvature state after every input limits the information available to its updates. Finally, PASLE provides modest but consistent gains via selective pseudo-labeling, while MEMO's full-model updates can be unstable when guided by a single unlabeled input. Overall, ELaTTA achieves the highest average accuracy among the compared methods without backpropagation.

\begin{table*}[t]
\caption{Accuracy (\%) on ImageNet-C with ViT-Base under strict single-instance adaptation. GF denotes gradient-free adaptation. Bold marks the best value in each column.}
\label{tab:ic_1}
\centering
\resizebox{\textwidth}{!}{
\begin{tabular}{l c ccc cccc cccc cccc >{\columncolor{gray!20}}c}
\toprule
\multirow{2}{*}{Method} & \multirow{2}{*}{GF} & \multicolumn{3}{c}{Noise} & \multicolumn{4}{c}{Blur} & \multicolumn{4}{c}{Weather} & \multicolumn{4}{c}{Digital} & Average\\ 
& & Gauss. & Shot & Impl. & Defoc. & Glass & Motion & Zoom & Snow & Frost & Fog & Brit. & Contr. & Elas. & Pix. & JPEG & Acc. \\ 
\midrule 
No Adapt          & \ding{51} & 55.34 & 56.23 & 56.01 & 46.48 & 34.78 & 52.87 & 44.20 & 62.39 & 62.66 & 65.56 & 77.70 & 32.04 & 45.73 & 66.72 & 66.67 & 55.03 \\ 
FOA   & \ding{51} & 53.87 & 54.16 & 54.00 & 46.17 & 33.45 & 52.56 & 43.69 & 61.82 & 62.30 & 66.17 & 77.73 & 30.60 & 46.14 & 66.18 & 66.77 & 54.37 \\ 
CAZO  & \ding{51} & 54.24 & 56.10 & 55.95 & 47.13 & 36.08 & 53.55 & 45.26 & 62.99 & 61.97 & 66.00 & 77.24 & 33.08 & 47.25 & 67.16 & 67.03 & 55.40 \\
T3A   & \ding{51} & 54.69 & 55.95 & 55.61 & 47.41 & 36.77 & 53.91 & \textbf{46.44} & 63.85 & 60.42 & 68.12 & 78.11 & 37.79 & \textbf{49.54} & 67.24 & 68.04 & 56.26 \\
CoTTA & \ding{55} & 54.61 & 55.66 & 55.37 & 45.28 & 34.35 & 52.69 & 44.11 & 62.38 & 62.62 & 58.33 & 77.71 & 29.58 & 45.65 & 66.68 & 66.66 & 54.11 \\
SAR   & \ding{55} & 55.25 & 56.08 & 55.89 & 46.22 & 34.41 & 52.28 & 43.82 & 62.09 & 62.69 & 65.56 & 77.53 & 32.03 & 45.47 & 66.37 & 66.55 & 54.82 \\
PASLE & \ding{55} & 56.72 & 56.24 & 56.21 & 47.53 & 35.32 & 53.02 & 44.03 & 62.43 & 62.81 & 65.84 & 78.62 & 31.23 & 46.65 & 66.76 & 67.24 & 55.38 \\
BECoTTA & \ding{55} & 55.67 & 56.45 & 56.29 & 46.21 & 33.68 & 52.66 & 43.67 & 62.20 & 63.37 & \textbf{68.25} & 77.58 & 33.74 & 45.09 & 66.70 & 66.78 & 55.22 \\
SURGEON & \ding{55} & 58.70 & 59.22 & 59.23 & 48.82 & 35.29 & 55.06 & 45.87 & 64.83 & 65.94 & 61.76 & 79.56 & 34.46 & 46.90 & 69.02 & 69.36 & 56.93 \\
MEMO & \ding{55} & 55.90 & 54.20 & 56.30 & 45.79 & \textbf{39.34} & 53.02 & 45.13 & 42.56 & 47.82 & 65.31 & 80.01 & \textbf{69.63} & 49.21 & 69.51 & \textbf{71.33} & 56.34 \\
\midrule
ELaTTA & \ding{51} & \textbf{58.77} & \textbf{59.66} & \textbf{59.50} & \textbf{49.30} & 36.08 & \textbf{55.35} & 46.34 & \textbf{65.21} & \textbf{66.40} & 67.66 & \textbf{80.21} & 35.96 & 47.61 & \textbf{69.55} & 69.68 & \textbf{57.82} \\
\bottomrule
\end{tabular}
}
\end{table*}

\noindent\textbf{Natural and appearance shifts.}
On ImageNet-V2, ImageNet-R, and ImageNet-Sketch, ELaTTA achieves an average accuracy of 63.72\%, compared with 59.96\% for No Adapt (\Tref{tab:ic_2}). CAZO reaches 60.37\%, again yielding a smaller gain in the strict protocol. ELaTTA gives the best results on ImageNet-V2 and ImageNet-R. On ImageNet-Sketch, it improves over No Adapt by 2.84 points, while T3A achieves the highest accuracy at 48.44\%.

\begin{table}[t]
\caption{Accuracy (\%) on ImageNet-V2/R/Sketch with ViT-Base under strict single-instance adaptation. GF denotes gradient-free adaptation. Bold marks the best value in each column.}
\label{tab:ic_2}
\centering
\begin{tabular}{l c c c c >{\columncolor{gray!20}}c}
\toprule
\multirow{2}{*}{Method} & \multirow{2}{*}{GF} & \multicolumn{4}{c}{Accuracy (\%)} \\
& & V2 & R & Sketch & Avg. \\
\midrule
No Adapt          & \ding{51} & 75.49 & 59.49 & 44.89 & 59.96 \\
FOA  & \ding{51} & 75.25 & 59.96 & 44.95 & 60.05 \\
CAZO & \ding{51} & 75.43 & 59.78 & 45.91 & 60.37 \\
T3A & \ding{51} & 75.61 & 57.98 & \textbf{48.44} & 60.68 \\
CoTTA & \ding{55} & 75.50 & 59.20 & 44.77 & 59.82 \\
SAR & \ding{55} & 75.33 & 59.39 & 44.82 & 59.85 \\
PASLE& \ding{55} & 75.66 & 61.73 & 45.72 & 61.04 \\
MEMO & \ding{55} & 76.08 & 62.85 & 46.08 & 61.67 \\
\midrule
ELaTTA         & \ding{51} & \textbf{78.15} & \textbf{65.29} & 47.73 & \textbf{63.72} \\
\bottomrule
\end{tabular}
\end{table}

\subsection{Efficiency and Edge Deployment}
\label{sec:computation}
\textbf{Compute, memory, and latency.}
As shown in \Fref{fig:comparison} and \Tref{tab:com}, ELaTTA demonstrates significant advantages in computational complexity compared to other methods. Specifically, ELaTTA achieves a GFLOPs value of 16.95, which is among the lowest across all methods, highlighting its high computational efficiency. T3A suffers from longer runtime despite having the lowest GFLOPs, due to its computation being concentrated in the final linear layer and entropy filtering step during support set updates. In terms of memory usage, ELaTTA requires only 696 MB, making it the most memory-efficient approach in the comparison. Moreover, ELaTTA achieves a short runtime per sample, at just 0.042 seconds, significantly outperforming other methods such as MEMO (1.009 s) and CoTTA (0.703 s). FOA has similiar peak memory, but ELaTTA is much faster. CAZO also avoids backpropagation, yet its adapter perturbations require repeated forward evaluations through the network. It uses 691.47 GFLOPs and 0.382 s per sample, compared with ELaTTA's 16.95 GFLOPs and 0.042 s. ELaTTA encodes each image once and evaluates candidates using the cached latent and classifier, while FOA evaluates input prompts through the encoder. Gradient-based SAR achieve slightly shorter running time by only updating the affine parameters in normalization layers, thereby reducing the computational cost of parameter updates. However, as shown in \Tref{tab:ic_1}, this strategy struggles in single-sample scenarios, where updating affine parameters alone may not be sufficient to achieve effective TTA.

\begin{table}[t]
\caption{Computational comparison (GFLOPs, Memory, and Latency) on ImageNet-C with ViT-Base.}
\label{tab:com}
\centering
\begin{tabular}{l c c c c}
\toprule
Method & GF & GFLOPs & Mem (MB) & Latency (s) \\
\midrule
FOA     & \ding{51} & 479.31 & 702 & 0.273 \\
CAZO    & \ding{51} & 691.47 & 767 & 0.382 \\
T3A     & \ding{51} & \textbf{16.86} & 718 & 0.124 \\
CoTTA   & \ding{55} & 50.59 & 1130 & 0.703 \\
SAR     & \ding{55} & 33.73 & 2996 & \textbf{0.037} \\
PASLE   & \ding{55} & 607.13 & 2588 & 0.051 \\
MEMO    & \ding{55} & 1096.14 & 8632 & 1.009 \\
BECoTTA & \ding{55} & 62.74 & 778 & 0.082 \\
SURGEON & \ding{55} & 26.24 & 716 & 0.071 \\
\midrule
ELaTTA  & \ding{51} & 16.95 & \textbf{696} & 0.042 \\
\bottomrule
\end{tabular}
\end{table}

\noindent\textbf{Raspberry Pi ViT deployment.}
We verify on-device feasibility on a Raspberry Pi 5 with 8~GB RAM, using PyTorch on the CPU with 4 threads and batch size one. No Adapt and ELaTTA load the same ViT-Base checkpoint in FP32 and share identical preprocessing and the ImageNet-C severity-5 evaluation protocol. ELaTTA runs strict single-instance adaptation with $k=16$ and $n=8$, resetting the search state for each image. \Tref{tab:rpi_vit} reports accuracy, encoder and search latency, and peak process RSS. The encoder runs once per image for both methods. One encoder forward pass takes 652~ms, while the isolated ELaTTA search takes 87~ms, or 13.3\% of the encoder latency. The complete ELaTTA pipeline takes 754~ms, compared with 663~ms for No Adapt, an end-to-end increase of 91~ms. The search evaluates $E=\lambda n=96$ candidates per image, with $\lambda=\lfloor 4+3\ln(16)\rfloor=12$ and $n=8$. Each candidate is a classifier pass on the cached latent, so the encoder still runs once per image. The Raspberry Pi accuracies are within 0.12 percentage points of the corresponding GPU results, indicating negligible numerical change after transfer to the device. The persistent deployment-specific basis $\mathbf{V}_k$ occupies approximately 0.05~MB, and no adaptation state is carried between images under the strict protocol. Peak process RSS increases from 1081~MB to 1109~MB, a 28~MB increase during adaptation.

\begin{table*}[t]
\centering
\caption{ViT-Base deployment on Raspberry Pi 5 (8~GB) using PyTorch CPU in FP32 with batch size one. Accuracy is measured on ImageNet-C at severity 5. The encoder runs once per image for both methods. ELaTTA search latency includes latent updates, classifier/entropy evaluations, and CMA-ES sampling and state updates; search is not applicable to No Adapt. Peak process RSS denotes the maximum resident set size of the inference process.}
\label{tab:rpi_vit}
\resizebox{\textwidth}{!}{%
\begin{tabular}{lccccc}
\toprule
Method & Accuracy (\%) & Encoder latency (ms) & Search latency (ms) & End-to-end latency (ms) & Peak process RSS (MB) \\
\midrule
No Adapt & 55.02 & 652 & N/A & 663 & 1081 \\
ELaTTA ($k=16$, $n=8$) & 57.94 & 652 & 87 & 754 & 1109 \\
\bottomrule
\end{tabular}}
\end{table*}

\subsection{Continual Adaptation}
\label{sec:generalization}
ELaTTA initializes each search from the previous coefficient in continual single-instance evaluation. In continual-batch evaluation, it searches one coefficient shared by 64 inputs and uses the batch solution to initialize the next batch. Both modes retain a $k$-dimensional vector between searches. The continual comparisons also include ROID \citep{roid}, ZOA \citep{zoa}, and Balance \citep{yang2024versatile}. All methods process the same input stream, with corruptions in the order displayed in the tables and no adaptation-state reset at corruption boundaries.

\noindent\textbf{Continual single-instance adaptation.}
ELaTTA achieves the highest average accuracy of 59.79\%, improving over No Adapt by 4.76 points and on 14 of the 15 corruptions (\Tref{tab:con-s}). CAZO reaches 58.14\%, a 2.74-point increase over its strict result, retaining adapter and curvature state across inputs appears to help, although single-input updates remain less effective than ELaTTA's latent search. CoTTA, ROID, and ZOA fall to near-chance accuracy, while FOA also degrades markedly. This pattern is consistent with error accumulation in continual adaptation: unreliable predictions can bias an update, and the resulting state influences subsequent inputs. With one sample per update, sample-specific errors are not averaged across a batch, making this feedback particularly difficult to control. FOA additionally relies on test-feature statistics for its search objective, which estimates from a single input provide a weak basis for distribution alignment. SAR remains competitive at 59.47\%, consistent with its use of reliable-sample selection and sharpness-aware updates to control noisy adaptation signals.

ELaTTA carries adaptation forward through a $k$-dimensional coefficient. Each new input is encoded by the same source model, and the inherited coefficient is reoptimized for that input within the fixed latent subspace. This confines temporal transfer to a feature correction without accumulating changes in the backbone weights. The 1.97-point gain over strict ELaTTA suggests that warm starts reuse useful corrections within corruption segments under the fixed search budget.

\begin{table*}[ht]
\caption{Accuracy (\%) on ImageNet-C with ViT-Base under continual single-instance adaptation. Corruptions follow the displayed stream order. Bold marks column maxima.}
\label{tab:con-s}
\centering
\resizebox{\textwidth}{!}{
\begin{tabular}{l ccc cccc cccc cccc >{\columncolor{gray!20}}c}
\toprule
\multirow{3}{*}{Method} &  \multicolumn{16}{l}{Time $ \xrightarrow{\hspace*{19cm}}$} \\
& \multicolumn{3}{c}{Noise} & \multicolumn{4}{c}{Blur} & \multicolumn{4}{c}{Weather} & \multicolumn{4}{c}{Digital} & Average\\ 
& Gauss. & Shot & Impl. & Defoc. & Glass & Motion & Zoom & Snow & Frost & Fog & Brit. & Contr. & Elas. & Pix. & JPEG & Acc. \\ 
\midrule 
No Adapt & 55.34 & 56.23 & 56.01 & 46.48 & 34.78 & 52.87 & 44.20 & 62.39 & 62.66 & 65.56 & 77.70 & 32.04 & 45.73 & 66.72 & 66.67 & 55.03 \\ 
ZOA & 0.06  & 0.06  & 0.06  & 0.06  & 0.06  & 0.06  & 0.06  & 0.06  & 0.06  & 0.06  & 0.06  & 0.06  & 0.06  & 0.06  & 0.06  & 0.06  \\
ROID & 0.10  & 0.08  & 0.08  & 0.08  & 0.08  & 0.08  & 0.08  & 0.08  & 0.08  & 0.08  & 0.08  & 0.08  & 0.08  & 0.08  & 0.08  & 0.08  \\
CoTTA & 0.13 & 0.12 & 0.16 & 0.12 & 0.13 & 0.16 & 0.13 & 0.16 & 0.14 & 0.17 & 0.21 & 0.11 & 0.15 & 0.17 & 0.15 & 0.15 \\
FOA & 6.94  & 1.74  & 1.42  & 1.72  & 0.40  & 0.48  & 0.68  & 0.86  & 0.72  & 0.98  & 2.04  & 0.88  & 0.86  & 0.22  & 0.96  & 1.39  \\
Balance & 3.86  & 6.68  & 1.82  & 0.13  & 9.16  & 8.10  & 3.77  & 0.82  & 0.34  & 0.14  & 79.24 & 0.10  & 0.21  & \textbf{73.60} & 70.95 & 17.26 \\
T3A & 55.18 & 56.72 & 56.00 & 38.88 & 32.96 & 50.96 & 42.82 & 60.14 & 60.18 & 64.22 & 76.48 & \textbf{40.24} & 43.12 & 66.60 & 68.48 & 54.20 \\
BECoTTA & 56.46 & 55.86 & 55.90 & 47.60 & 34.80 & 52.34 & 43.34 & 63.30 & 62.26 & 66.34 & 77.88 & 31.60 & 45.64 & 66.08 & 66.36 & 55.05 \\
CAZO & 60.65 & 59.83 & 61.41 & 46.82 & 39.64 & 57.98 & 45.30 & 64.33 & 66.48 & 66.84 & 78.00 & 38.42 & 50.30 & 68.63 & 67.45 & 58.14 \\
SAR & 59.08 & 60.52 & 59.36 & 45.52 & \textbf{57.26} & \textbf{58.56} & \textbf{57.12} & 62.74 & 66.66 & 68.68 & 78.78 & 6.68  & \textbf{67.16} & 72.40 & \textbf{71.56} & 59.47 \\
\midrule
ELaTTA & \textbf{61.78} & \textbf{62.40} & \textbf{62.98} & \textbf{51.82} & 39.50 & 57.74 & 47.44 & \textbf{68.92} & \textbf{68.52} & \textbf{68.72} & \textbf{80.80} & 31.74 & 56.20 & 70.30 & 67.94 & \textbf{59.79} \\
\bottomrule
\end{tabular}}
\end{table*}

\noindent\textbf{Continual-batch adaptation.}
With batch size 64, ELaTTA reaches 59.53\% average accuracy, improving over No Adapt by 4.50 points. CAZO achieves the highest average accuracy at 63.21\%, followed by FOA, ROID, and T3A at 61.60\%, 60.93\%, and 60.49\%, respectively (\Tref{tab:con-b}). These methods differ in where and how they adapt. CAZO's adapter updates can alter features within the encoder, and the larger batch provides more context for its feature-statistic objective and zeroth-order estimates. This is consistent with its stronger result here than in either single-instance protocol. FOA optimizes input prompts using entropy and source--test feature-statistic alignment, while ROID updates normalization parameters within the network. Both can change representations during feature extraction. ELaTTA applies a shared additive correction after encoding, restricted to a fixed 16-dimensional subspace. It cannot adjust the encoder or correct directions outside that subspace, and the same coefficient must serve all 64 inputs. The shared correction translates batch features without changing their pairwise differences, which may limit its benefit from batch feedback. FOA's 32.00-point lead on contrast accounts for much of their average gap.

The restriction also limits computation: ELaTTA encodes each image once and reuses its latent for every candidate. FOA's prompt search and CAZO's adapter updates require repeated network evaluations, while ROID's normalization updates require backpropagation. At batch size 64, ELaTTA uses the least memory at 1,290 MB, close to FOA's 1,300 MB and below CAZO's 1,763 MB. CAZO's average accuracy is 3.68 points higher. Compared with ROID, ELaTTA reduces memory by 88.1\% with a 1.40-point accuracy reduction; compared with T3A, it reduces memory by 81.1\% with a 0.96-point reduction.

\begin{table*}[ht]
\caption{Accuracy (\%) and memory use (MB) on ImageNet-C with ViT-Base under continual-batch adaptation (batch size 64). Bold marks the highest accuracies and the lowest reported memory.}
\label{tab:con-b}
\centering
\resizebox{\textwidth}{!}{
\begin{tabular}{l ccc cccc cccc cccc >{\columncolor{gray!20}}c r}
\toprule
\multirow{3}{*}{Method} & \multicolumn{16}{l}{Time $ \xrightarrow{\hspace*{19cm}}$} & \multirow{3}{*}{\shortstack{Memory\\(MB)}} \\
& \multicolumn{3}{c}{Noise} & \multicolumn{4}{c}{Blur} & \multicolumn{4}{c}{Weather} & \multicolumn{4}{c}{Digital} & Average & \\
& Gauss. & Shot & Impl. & Defoc. & Glass & Motion & Zoom & Snow & Frost & Fog & Brit. & Contr. & Elas. & Pix. & JPEG & Acc. & \\
\midrule 
No Adapt & 55.34 & 56.23 & 56.01 & 46.48 & 34.78 & 52.87 & 44.20 & 62.39 & 62.66 & 65.56 & 77.70 & 32.04 & 45.73 & 66.72 & 66.67 & 55.03 & -- \\
ZOA & 58.00 & 58.56 & 59.38 & 47.98 & 39.10 & 56.38 & 49.38 & 65.68 & 63.48 & 62.68 & 79.34 & 49.02 & 50.76 & 66.84 & 70.34 & 58.46 & 1734 \\
ROID & 60.68 & 63.04 & 63.00 & \textbf{56.60} & 55.84 & \textbf{62.72} & \textbf{58.86} & 66.78 & 66.70 & 70.76 & 79.24 & 12.14 & \textbf{57.02} & 69.46 & 71.04 & 60.93 & 10866 \\
FOA & 57.14 & 61.62 & 61.96 & 51.98 & 42.78 & 57.48 & 54.08 & 65.98 & 68.44 & 65.60 & 79.48 & \textbf{62.78} & 52.20 & 70.32 & \textbf{72.20} & 61.60 & 1300 \\
Balance & 58.52 & \textbf{63.92} & \textbf{64.46} & 50.34 & 40.66 & 56.58 & 44.94 & 67.62 & 68.42 & 66.48 & 77.74 & 39.54 & 51.66 & 68.46 & 69.70 & 59.27 & 19489 \\
T3A & 58.70 & 61.18 & 61.40 & 47.48 & 48.88 & 58.38 & 50.92 & 62.82 & 62.30 & 64.90 & 78.32 & 60.40 & 54.78 & 67.50 & 69.44 & 60.49 & 6818 \\
BECoTTA & \textbf{61.82} & 62.52 & 62.27 & 51.64 & 38.10 & 58.16 & 48.65 & 67.79 & \textbf{68.82} & 36.77 & \textbf{82.20} & 30.69 & 49.38 & \textbf{71.99} & 72.11 & 57.53 & 3340 \\
CAZO & 60.65 & 63.29 & 62.13 & 54.26 & \textbf{55.86} & 61.02 & 55.19 & 66.26 & 67.76 & 69.07 & 79.95 & 60.83 & 55.08 & 69.80 & 66.98 & \textbf{63.21} & 1763 \\
SAR & 55.18 & 56.72 & 56.02 & 38.76 & 33.04 & 50.98 & 42.72 & 60.12 & 60.12 & 64.22 & 76.48 & 40.22 & 43.10 & 66.62 & 68.44 & 54.18 & 1458 \\
\midrule
ELaTTA & 61.38 & 61.98 & 62.64 & 51.08 & 39.16 & 57.70 & 47.20 & \textbf{68.76} & 68.60 & \textbf{71.50} & 79.96 & 30.78 & 55.78 & 69.28 & 67.08 & 59.53 & \textbf{1290} \\
\bottomrule
\end{tabular}}
\end{table*}

\subsection{Architecture and Domain-Shift Generalization}
\label{sec:architecture_generalization}
\textbf{Visual backbones.}
We evaluate ELaTTA across ResNet-50 \citep{he2016deep}, EfficientNet-B0 \citep{tan2019efficientnet}, MobileNet-V4 \citep{qin2024mobilenetv4}. As all use BN, we include BN-specific baselines IABN \citep{gong2022note} and TTN \citep{lim2023ttn} in \Tref{tab:diverse_networks}. On ResNet-50, ELaTTA improves accuracy on most domains, yielding an average gain of 2.83 percentage points. It also boosts the lightweight EfficientNet-B0 (5.29M) and MobileNet-V4 (3.77M), though gains are limited under extreme corruptions (e.g., Glass Blur on MobileNet-V4, Contrast on ResNet-50) where accuracy is very low (3.03\%, 0.72\%). This likely reflects catastrophic feature degradation. When semantic signal vanishes, latent-space TTA becomes ineffective, suggesting stronger backbone robustness is required in such regimes. This limitation is further examined in the corruption severity analysis in \Sref{sec:abl}.

\begin{table*}[t]
\caption{Accuracy (\%) on ImageNet-C across visual backbones under strict single-instance adaptation. Bold marks the best value for each backbone and corruption.}
\label{tab:diverse_networks}
\centering
\resizebox{\textwidth}{!}{
\begin{tabular}{l c ccc cccc cccc cccc >{\columncolor{gray!20}}c}
\toprule
\multirow{2}{*}{Networks} & \multirow{2}{*}{Method} & \multicolumn{3}{c}{Noise} & \multicolumn{4}{c}{Blur} & \multicolumn{4}{c}{Weather} & \multicolumn{4}{c}{Digital} & Average\\ 
& & Gauss. & Shot & Impl. & Defoc. & Glass & Motion & Zoom & Snow & Frost & Fog & Brit. & Contr. & Elas. & Pix. & JPEG & Acc. \\ 
\midrule 
\multirow{5}{*}{ResNet-50} & No Adapt & 4.47 & 4.74 & 4.06 & 8.11 & 5.96 & 9.59 & 14.74 & 6.92 & 14.47 & 12.32 & 45.80 & 0.72 & 11.08 & 18.19 & 32.54 & 12.91 \\ 
 & IABN     & \textbf{7.22} & 5.10 & \textbf{5.54} & 5.46 & \textbf{7.38} & 11.92 & 15.98 & 5.74 & 12.32 & 14.14 & 47.10 & \textbf{0.90} & 7.10 & 18.26 & 18.04 & 12.15 \\
 & TTN      & 1.73 & 5.13 & 4.90 & 4.17 & 6.69 & 10.16 & 14.14 & 6.13 & 14.52 & 14.42 & 49.20 & 0.22 & 12.26 & 22.03 & 33.07 & 13.25 \\
 & SAR      & 2.34 & 4.46 & 4.24 & 6.56 & 6.66 & 14.04 & 16.34 & 4.02 & 14.68 & 13.26 & 48.51 & 0.35 & 12.43 & 20.44 & 36.46 & 13.65 \\
& ELaTTA & 4.99 & \textbf{5.18} & 4.41 & \textbf{11.36} & 7.32 & \textbf{14.26} & \textbf{18.29} & \textbf{7.00} & \textbf{15.32} & \textbf{15.01} & \textbf{55.32} & 0.25 & \textbf{13.07} & \textbf{23.29} & \textbf{40.97} & \textbf{15.74} \\ 
\midrule 
\multirow{5}{*}{EfficientNet-B0} & No Adapt & 15.07 & 18.47 & 14.66 & 21.48 & 8.66 & 21.73 & 24.24 & 30.95 & 27.77 & 28.80 & 67.25 & 21.96 & 17.62 & 46.73 & 50.51 & 27.73 \\
& IABN     & 14.71 & 19.32 & 14.75 & 21.34 & \textbf{9.24} & 21.59 & 25.65 & 33.35 & 29.45 & 29.56 & 69.24 & 22.51 & 17.88 & 49.46 & 53.32 & 28.76 \\
& TTN      & 15.32 & 12.34 & 15.32 & 23.44 & 6.32  & 23.23 & 26.36 & 32.02 & 28.66 & 32.74 & 69.35 & 24.78 & \textbf{18.50} & 52.64 & 54.44 & 29.03 \\
& SAR      & 15.34 & 20.58 & 16.22 & 22.53 & 4.24  & 20.34 & 22.35 & 32.45 & 20.76 & 28.63 & 68.64 & 22.46 & 16.34 & 46.86 & 49.58 & 27.15 \\
& ELaTTA & \textbf{16.61} & \textbf{21.53} & \textbf{16.71} & \textbf{25.02} & 7.57 & \textbf{25.91} & \textbf{27.69} & \textbf{35.05} & \textbf{30.38} & \textbf{32.78} & \textbf{75.67} & \textbf{25.96} & \textbf{18.50} & \textbf{55.14} & \textbf{59.24} & \textbf{31.58} \\ 
\midrule 
\multirow{5}{*}{MobileNet-V4} & No Adapt & 5.97 & \textbf{7.58} & 6.02 & 9.69 & 3.03 & 12.30 & 11.19 & 12.57 & 20.20 & 11.41 & 58.11 & \textbf{3.84} & 11.05 & 10.29 & 32.79 & 14.40 \\ 
& IABN     & 5.32 & 7.50 & 6.23 & 9.46 & 3.02 & 12.34 & 11.68 & 12.34 & 20.68 & \textbf{13.46} & 66.48 & 2.02 & 11.36 & 9.68  & 36.46 & 15.20 \\
& TTN      & \textbf{6.98} & 7.54 & \textbf{6.38} & 9.84 & \textbf{3.34} & 13.56 & \textbf{12.56} & 12.64 & 20.46 & 12.42 & 64.66 & 2.46 & 10.48 & 10.46 & 40.58 & 15.62 \\
& SAR      & 4.98 & 7.32 & 4.54 & 10.20& 3.02 & 14.54 & 11.08 & 12.78 & 22.34 & 11.60 & 60.34 & 2.66 & 10.46 & 11.42 & 39.64 & 15.13 \\
& ELaTTA & 5.47 & 7.55 & 5.43 & \textbf{10.89} & 2.99 & \textbf{14.57} & 12.28 & \textbf{13.17} & \textbf{23.08} & 12.70 & \textbf{68.52} & 2.78 & \textbf{11.77} & \textbf{11.77} & \textbf{41.43} & \textbf{16.29} \\
\bottomrule
\end{tabular}
}
\end{table*}

\noindent\textbf{Source in-distribution performance.}
We evaluate ELaTTA on clean source test data under strict single-instance adaptation (\Tref{tab:id}). Accuracy improves across all four backbones, extending the benefits of latent correction to the source distribution. With the network weights and source-derived basis fixed, per-input coefficient search refines the predictions using the existing representations. These results suggest that the source principal subspace provides useful directions for prediction refinement even on clean inputs. Together with the corruption results, they support a unified adaptation procedure for clean and shifted inputs without a separate shift-detection stage.

\begin{table}[t]
\centering
\newsavebox{\tabidbox}
\sbox{\tabidbox}{%
\resizebox{\columnwidth}{!}{%
\begin{tabular}{l c c c c}
\toprule
Model & Vit-Base & ResNet-50 & EfficientNet-B0 & MobileNet-V4 \\
\midrule
No Adapt  & 85.16 & 70.74 & 78.53 & 71.04 \\
ELaTTA       & 87.05 & 76.71 & 85.31 & 79.91 \\
\midrule
\textbf{Improvement} & \textbf{+1.89} & \textbf{+5.97} & \textbf{+6.78} & \textbf{+8.87} \\
\bottomrule
\end{tabular}
}%
}
\caption{Accuracy (\%) on clean source test data under strict single-instance adaptation. Improvements are reported in percentage points.}
\label{tab:id}
\usebox{\tabidbox}
\end{table}

\noindent\textbf{DomainNet-126.}
To further assess the robustness of ELaTTA under rigorous conditions beyond synthetic corruptions (e.g., ImageNet-C), we extend our continual evaluation to DomainNet-126 \citep{peng2019moment}. Unlike corruption benchmarks that primarily introduce texture or noise degradations while preserving object geometry, DomainNet features significant semantic and stylistic variations across four distinct domains (Real, Sketch, Clipart, and Painting), presenting a substantial challenge for adaptation methods.

As shown in \Tref{tab:domainnet126}, on DomainNet-126, ELaTTA remains competitive in continual-batch adaptation and achieves the highest average accuracy among the compared methods in continual single-instance adaptation for all four source domains. The single-instance baselines show substantial performance degradation, whereas ELaTTA maintains gains over No Adapt across all source-domain groups. These results extend its effectiveness beyond synthetic corruptions to cross-domain appearance and style shifts.

\begin{table*}[t]
\centering
\caption{Accuracy (\%) on DomainNet-126 with ResNet-50. Each group specifies the source domain and its three target domains. BS denotes test-time batch size.}
\label{tab:domainnet126}
\resizebox{\textwidth}{!}{
\begin{tabular}{l | ccc >{\columncolor{gray!20}}c | ccc >{\columncolor{gray!20}}c | ccc >{\columncolor{gray!20}}c | ccc >{\columncolor{gray!20}}c}
\toprule
\multirow{2}{*}{Method} & \multicolumn{4}{c|}{(a) \texttt{Painting} as Source} & \multicolumn{4}{c|}{(b) \texttt{Clipart} as Source} & \multicolumn{4}{c|}{(c) \texttt{Real} as Source} & \multicolumn{4}{c}{(d) \texttt{Sketch} as Source} \\
\cmidrule(lr){2-5} \cmidrule(lr){6-9} \cmidrule(lr){10-13} \cmidrule(l){14-17}
& Real & Sketch & Clipart & Avg. & Sketch & Real & Painting & Avg. & Clipart & Painting & Sketch & Avg. & Painting & Clipart & Real & Avg. \\
\midrule
No Adapt     & 74.84 & 49.70 & 53.26 & 59.27 & 46.16 & 60.50 & 43.84 & 50.17 & 55.62 & 61.88 & 47.84 & 55.11 & 48.48 & 54.88 & 59.32 & 54.23 \\
SAR (BS=64)  & 73.58 & 53.96 & 53.50 & 60.35 & 48.80 & 64.36 & 47.34 & 53.50 & 54.36 & 62.90 & 48.40 & 55.22 & 57.56 & 58.98 & 67.92 & 61.49 \\
ROID (BS=64) & 75.04 & 57.30 & 57.24 & 63.19 & 51.24 & 64.64 & 49.14 & 55.01 & 57.48 & 65.18 & 52.66 & 58.44 & 58.76 & 61.44 & 68.58 & 62.93 \\
SAR (BS=1)   & 6.42  & 4.24  & 3.22  & 4.63  & 4.98  & 5.80  & 4.38  & 5.05  & 3.74  & 4.66  & 3.58  & 3.99  & 4.78  & 4.50  & 5.26  & 4.85  \\
ROID (BS=1)  & 0.48  & 0.16  & 0.16  & 0.27  & 0.16  & 0.48  & 0.30  & 0.31  & 0.16  & 0.30  & 0.16  & 0.21  & 0.30  & 0.16  & 0.48  & 0.31  \\
MEMO (BS=1)  & 40.38 & 17.82 & 25.42 & 27.87 & 40.38 & 15.48 & 27.92 & 27.93 & 40.38 & 26.12 & 22.92 & 29.81 & 13.92 & 26.00 & 21.68 & 20.53 \\
\midrule
ELaTTA (BS=64) & 78.24 & 53.74 & 55.50 & 62.49 & 47.24 & 63.24 & 46.08 & 52.19 & 56.40 & 63.28 & 49.82 & 56.50 & 56.52 & 61.32 & 64.46 & 60.77 \\
ELaTTA (BS=1)  & 76.08 & 52.94 & 54.22 & 61.08 & 47.20 & 62.44 & 44.74 & 51.46 & 56.50 & 63.12 & 49.16 & 56.26 & 54.44 & 59.50 & 63.50 & 59.15 \\
\bottomrule
\end{tabular}
}
\end{table*}

\subsection{Cross-Modal Applicability to Keyword Spotting}
\label{sec:cross_modal}
\begin{table*}[t]
\caption{Performance comparison on GSC-C with LSTM model regarding \textbf{Accuracy} (\%). \textbf{GF} stands for gradient-free. The \textbf{bold} number indicates the best result.}
\label{tab:kws}
\centering
\resizebox{\textwidth}{!}{
\begin{tabular}{l c c cc cc cc cc cc >{\columncolor{gray!20}}c}
\toprule
\multirow{2}{*}{SNR} & \multirow{2}{*}{Method} & \multirow{2}{*}{GF} & \multicolumn{2}{c}{Animals} & \multicolumn{2}{c}{Natural} & \multicolumn{2}{c}{Human} & \multicolumn{2}{c}{Domestic} & \multicolumn{2}{c}{Urban} & Average\\ 
& & & dog & cat & pouring water & thunderstorm & crying baby & laughing & washing machine & vacuum cleaner & car horn & fireworks & Acc. \\
\midrule 
\multirow{4}{*}{-10 dB} & No Adapt    & \ding{51} & 62.67 & 61.17 & 54.55 & 66.23 & 58.74 & 58.59 & 52.88 & 50.43 & 56.62 & 61.54 & 58.34 \\
& T3A  & \ding{51} & 62.67 & 61.17 & 54.55 & 66.23 & 58.74 & 58.59 & 52.88 & 50.43 & 56.62 & 61.54 & 58.34 \\
& PASLE  & \ding{55} & 62.87 & 62.06 & 54.78 & \textbf{66.75} & 59.42 & 59.63 & 57.69 & 53.38 & 58.35 & 62.73 & 59.77 \\
& ELaTTA  & \ding{51} & \textbf{64.25} & \textbf{63.58} & \textbf{59.73} & 66.47 & \textbf{61.99} & \textbf{61.94} & \textbf{59.46} & \textbf{56.98} & \textbf{59.32} & \textbf{64.90} & \textbf{61.86} \\
\midrule
\multirow{4}{*}{-15 dB} & No Adapt    & \ding{51} & 57.08 & 53.63 & 49.35 & 61.45 & 53.08 & 53.03 & 46.81 & 47.49 & 51.76 & 55.19 & 52.89 \\
& T3A  & \ding{51} & 57.08 & 53.63 & 49.35 & 61.45 & 53.08 & 53.03 & 46.81 & 47.49 & 51.76 & 55.19 & 52.89 \\
& PASLE  & \ding{55} & 58.12 & 53.77 & 52.49 & 61.93 & 55.64 & 56.72 & 48.34 & 49.23 & 52.67 & 59.32 & 54.82 \\
& ELaTTA  & \ding{51} & \textbf{60.84} & \textbf{57.99} & \textbf{57.71} & \textbf{62.28} & \textbf{58.04} & \textbf{58.98} & \textbf{57.83} & \textbf{55.38} & \textbf{56.00} & \textbf{60.41} & \textbf{58.55} \\
\midrule 
\multirow{4}{*}{-20 dB} & No Adapt    & \ding{51} & 52.75 & 48.50 & 46.21 & 58.08 & 51.12 & 48.53 & 45.55 & 46.05 & 48.81 & 50.94 & 49.65 \\
& T3A & \ding{51} & 52.75 & 48.50 & 46.21 & 58.08 & 51.12 & 48.53 & 45.55 & 46.05 & 48.81 & 50.94 & 49.65 \\
& PASLE  & \ding{55} & 53.76 & 51.77 & 48.84 & 59.03 & 53.31 & 50.38 & 46.21 & 50.21 & 52.17 & 53.56 & 51.92 \\
& ELaTTA & \ding{51} & \textbf{59.07} & \textbf{54.71} & \textbf{57.20} & \textbf{59.35} & \textbf{56.94} & \textbf{57.94} & \textbf{58.22} & \textbf{55.54} & \textbf{54.50} & \textbf{58.07} & \textbf{57.15} \\
\bottomrule
\end{tabular}
}
\end{table*}

We apply ELaTTA to an LSTM trained on clean Google Speech Commands \citep{warden2018speech} under the strict single-instance setting. GSC-C mixes speech with ten background sounds from ESC-50 \citep{piczak2015esc}, covering animal, natural, human, domestic, and urban sounds, at SNRs of $-10$, $-15$, and $-20$ dB. For each utterance, we randomly crop a background-noise segment to match its duration and add it to the speech signal at the target SNR, preserving the original keyword label.

We omit CoTTA and MEMO because their speech-time counterparts require additional design choices, and FOA because its prompt-based formulation is not compatible with our LSTM KWS backbone. We omit CAZO because its ViT adapters and visual feature-alignment objective would require an LSTM-specific design. SAR is also excluded because our LSTM has no normalization layers whose affine parameters it can adapt. We therefore compare ELaTTA with T3A and PASLE on GSC-C, as shown in \Tref{tab:kws}. ELaTTA consistently achieves the best accuracy, with gains increasing as SNR decreases, clearly indicating that the fixed principal subspace ($\mathbf{V}_k$) provides effective source-domain guidance under heavier noise. In contrast, T3A is close to No Adapt due to insufficient confident per-class supports in the 12-class, single-instance setting. PASLE yields small but consistent improvements, further underscoring the practical advantage of ELaTTA for noisy KWS in low-SNR conditions.

\subsection{Ablation Analysis}
\label{sec:abl}
\begin{figure}[t]
\centering
\includegraphics[width=\columnwidth]{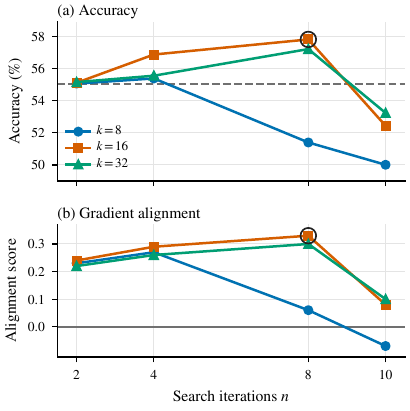}
\caption{Effect of the subspace dimension $k$ and the number of search iterations $n$ on ImageNet-C with ViT-Base under strict single-instance adaptation. (a) Average accuracy; the dashed line is No Adapt. (b) Alignment between the descent directions of the search objective and supervised cross-entropy. The circle marks the default setting.}
\label{fig:kn}
\end{figure}

\textbf{Subspace dimension and search iterations.}
\Fref{fig:kn}(a) reports accuracy for $k\in\{8,16,32\}$ and $n\in\{2,4,8,10\}$. With $n=2$ the search barely changes the prediction, and accuracy stays within 0.13 points of No Adapt for all three $k$. Accuracy then rises with $n$ to a peak, at $n=8$ for $k=16$ and $k=32$ and at $n=4$ for $k=8$, and falls beyond it. At $n=10$ all three settings are below No Adapt. Each setting costs $E=\lambda n$ candidate evaluations per image, so evaluations beyond the accuracy peak only add latency. The default setting takes 0.042 s per image (\Tref{tab:com}).

To examine the decline, we measure the alignment between the two loss-descent directions, $\cos(-\nabla_{\mathbf z} J_{\text{proxy}}, -\nabla_{\mathbf z} L_{\text{CE}})$, where $J_{\text{proxy}}$ is the search objective and $L_{\text{CE}}$ is the supervised cross-entropy on a labeled set used only for this analysis. Positive values indicate local agreement between the two descent directions, while negative values indicate opposition. \Fref{fig:kn}(b) shows that alignment follows accuracy. It peaks at the same $n$ and then decays, suggesting that the entropy proxy becomes less consistent with the supervised loss as the search proceeds beyond the accuracy peak. This is a local gradient diagnostic, rather than a measurement of the gradient-free CMA-ES update direction. The decay is faster for smaller $k$. From $n=4$ to $n=10$ the score drops by 0.34 for $k=8$, ending at $-0.07$, compared with 0.21 for $k=16$ and 0.16 for $k=32$. In these experiments, a small subspace therefore tolerates fewer iterations, and $k=8$ is already below No Adapt at $n=8$. The setting $k=16$, $n=8$ gives the highest accuracy, 57.82\%, and is used throughout.

\noindent\textbf{Source-cache size.}
\Tref{tab:source_N} compares the 50k-sample reference cache used in the main ImageNet experiments with smaller caches of $N\in\{30,1000,5000\}$ training images, keeping $k=16$ and $n=8$. For each smaller size, we draw five independent caches and report the mean and standard deviation. We measure subspace overlap as $\|\mathbf{V}_k^{(N)\top}\mathbf{V}_k^{(\mathrm{ref})}\|_F^2/k$, the mean squared cosine of the principal angles to the reference basis.

All cache sizes achieve mean accuracy within one point of the reference result (57.82\%), and $N=5$k is within 0.1 points. Increasing $N$ primarily improves consistency across cache draws: the accuracy standard deviation decreases from about one point at $N\leq 1$k to 0.16 points at $N=5$k, while overlap with the reference basis rises from 0.61 to 0.96. These results show that a small source cache can provide an effective adaptation subspace, while larger caches reduce sensitivity to which source samples are selected.

\begin{table}[t]
\centering
\newsavebox{\tabsourcenbox}
\sbox{\tabsourcenbox}{%
\resizebox{\columnwidth}{!}{%
\begin{tabular}{l cccc}
\toprule
$N$ & 30 & 1k & 5k & 50k \\
\midrule
Acc. (\%) & 58.69$\pm$1.02 & 58.12$\pm$0.94 & 57.89$\pm$0.16 & 57.82 \\
Overlap & 0.61$\pm$0.04 & 0.82$\pm$0.01 & 0.96$\pm$0.01 & 1.00 \\
\bottomrule
\end{tabular}
}}
\caption{Effect of the source-cache size $N$ on ImageNet-C with ViT-Base under strict single-instance adaptation ($k=16$, $n=8$). All caches are sampled from the ImageNet-1k training split. For $N\leq 5$k, values are the mean $\pm$ standard deviation over five independent caches. The fixed 50k-sample cache defines the reference basis for overlap.}
\label{tab:source_N}
\usebox{\tabsourcenbox}
\end{table}

\noindent\textbf{Calibration under distribution shift.}
\Tref{tab:calibration} reports expected calibration error (ECE) \citep{guo2017calibration} and NLL for No Adapt, backpropagation on the same subspace, and ELaTTA, pooled over the 15 corruptions at severity 5, with ECE computed over ten equal-width confidence bins. Both adaptation methods minimize entropy and thus raise prediction confidence, and both reduce ECE relative to No Adapt, from 8.64\% to 6.72\% and 6.23\%. They differ in NLL. Backpropagation improves accuracy by only 0.49 points and increases NLL from 2.32 to 2.34, indicating that part of the added confidence goes to incorrect predictions. ELaTTA improves accuracy by 2.79 points and reduces NLL to 2.03. The simultaneous improvements in accuracy, ECE, and NLL indicate better overall predictive performance and probability quality under the tested setting. The prediction-change analysis below provides complementary evidence of a net correction benefit.
\begin{table}[t]
\centering
\caption{Accuracy and calibration metrics on ImageNet-C with ViT-Base under strict single-instance adaptation. Accuracy and ECE are in \%.}
\label{tab:calibration}
\scriptsize
\begin{tabular}{lccc}
\toprule
Method & Acc $\uparrow$ & ECE $\downarrow$ & NLL $\downarrow$ \\
\midrule
No Adapt & 55.03 & 8.64	& 2.32 \\
Backprop. & 55.52	& 6.72 & 2.34 \\
ELaTTA & 57.82 & 6.23 & 2.03 \\
\bottomrule
\end{tabular}
\end{table}

\noindent\textbf{Optimizer comparison.}
To examine the search strategy, we compare CMA-ES with Backprop, (1+1)-Evolution Strategy ((1+1)-ES), and Zeroth-Order SGD (ZO-SGD) for latent coefficient adaptation. We report the best accuracy obtained in our hyperparameter search for each optimizer. CMA-ES achieves the highest accuracy in \Tref{tab:optimizer}, exceeding (1+1)-ES and Backprop by 2.09 and 2.30 points, respectively. The differences show that the optimizer affects how effectively the entropy signal is used for classification within the source subspace. CMA-ES evaluates several corrections per round and uses their ranks to adapt the location, shape, and scale of subsequent sampling. This feedback provides a practical way to explore the coefficient space using only classifier evaluations. The calibration results in \Tref{tab:calibration} also favor the selected CMA-ES configuration: it reduces NLL, whereas Backprop slightly increases it. The iteration study further shows that the search budget matters, with accuracy declining beyond its peak. Together, these results support CMA-ES with a finite evaluation budget for this adaptation task. Its forward-only operation and established Python and C/C++ implementations also make it practical for integration into edge-device runtimes.

\begin{table}[t]
\centering
\newsavebox{\taboptbox}
\sbox{\taboptbox}{%
\resizebox{\columnwidth}{!}{%
\begin{tabular}{l cccc >{\columncolor{gray!20}}c}
\toprule
Optimizer & (1+1) ES & ZO-SGD & Backprop & CMA-ES \\ 
\midrule 
Accuracy & 55.73 & 53.61 & 55.52 & 57.82 \\
\bottomrule
\end{tabular}}}
\caption{Accuracy (\%) on ImageNet-C with ViT-Base using different optimizers for latent coefficient adaptation.}
\label{tab:optimizer}
\usebox{\taboptbox}
\end{table}

\noindent\textbf{Contribution of the subspace and search strategy.}
\begin{table}[t]
\centering
\newsavebox{\tabcompabbox}
\sbox{\tabcompabbox}{%
\resizebox{\columnwidth}{!}{%
\begin{tabular}{lcc}
\toprule
Subspace & Search & Accuracy (\%) \\
\midrule
None & No Adapt & 55.03 \\
Random orthonormal & CMA-ES & 49.21 \\
Source principal & Fixed Gaussian & 54.12 \\
Source principal & CMA-ES & 57.82 \\
\bottomrule
\end{tabular}
}}
\caption{The subspace and search strategy ablation on ImageNet-C. Fixed Gaussian samples candidates from a non-adaptive Gaussian.}
\label{tab:component_ablation}
\usebox{\tabcompabbox}
\end{table}
The source basis and adaptive search contribute jointly to the gain in \Tref{tab:component_ablation}. All variants use the same candidate number, iterations, and initial search scale. Replacing the source basis with a random orthonormal basis lowers accuracy to 49.21\%, showing that the choice of correction directions matters even when their number is fixed. The source basis concentrates the search on variations represented in the source features. Within this basis, fixed Gaussian sampling reaches 54.12\%, while CMA-ES reaches 57.82\%. Both variants evaluate Gaussian samples and select the lowest-entropy candidate; CMA-ES additionally uses candidate feedback to adapt subsequent sampling. Their 3.70-point difference supports the value of this distribution update under the same evaluation budget. The full method improves over No Adapt by 2.79 points, showing the benefit of combining source principal directions with adaptive candidate sampling.

\noindent\textbf{Variation across search seeds.}
\begin{table}[t]
\centering
\caption{Variation across search seeds on ImageNet-C with ViT-Base. The source cache is fixed, and each entry is the mean $\pm$ standard deviation of the average accuracy over three independent search seeds.}
\label{tab:search_seeds}
\scriptsize
\begin{tabular}{lc}
\toprule
Protocol & Accuracy (\%) \\
\midrule
Strict single-instance & 57.86 $\pm$ 0.13 \\
Continual single-instance & 59.75 $\pm$ 0.11 \\
\bottomrule
\end{tabular}
\end{table}
The randomness of the CMA-ES search has little effect on the result (\Tref{tab:search_seeds}). With the source cache fixed, the standard deviation is about 0.1 points, more than twenty times smaller than the gain over No Adapt (2.8 points for strict and 4.7 points for continual adaptation). Continual adaptation shows similarly low variation in average accuracy across the tested search seeds, despite carrying the coefficient between inputs.

\noindent\textbf{Prediction changes.}
To examine how latent correction changes classification decisions, we track prediction correctness before and after adaptation. On Gaussian noise, ELaTTA corrects 2,161 initially incorrect predictions and changes 446 initially correct predictions to incorrect ones (\Tref{tab:switch_analysis}). Beneficial switches outnumber harmful switches by about 4.8 to one, yielding 1,715 additional correct predictions, or a 3.43-point accuracy gain across 50,000 inputs. Meanwhile, 27,224 predictions remain correct. These counts show that the net gain comes from correcting a subset of initial errors while retaining most initially correct decisions. The selected coefficients can therefore move inputs across decision boundaries in useful directions under the tested search budget. Together with the component and calibration results, this analysis supports the effectiveness of source-constrained adaptive search at both the decision and probability levels.

\begin{table}[t]
\caption{Prediction changes before and after ELaTTA on ImageNet-C Gaussian noise domain with ViT-Base model.}
\label{tab:switch_analysis}
\centering
\begin{tabular}{l r}
\toprule
Metric & Value \\ 
\midrule 
Correct $\rightarrow$ Correct & 27,224 \\
Incorrect $\rightarrow$ Incorrect & 20,169 \\
Beneficial Switch & 2,161 (4.32\%) \\
Harmful Switch & 446 (0.89\%) \\
\midrule
Correctness Transition Rate & 5.21\% \\
Net Gain & +1,715 \\
\bottomrule
\end{tabular}
\end{table}

\begin{figure}[t]
\centering
\includegraphics[width=\columnwidth]{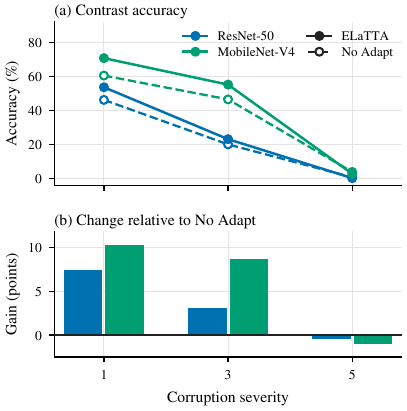}
\caption{Effect of corruption severity on the Contrast corruption of ImageNet-C under strict single-instance adaptation. (a) Accuracy of No Adapt (dashed) and ELaTTA (solid). (b) Accuracy change of ELaTTA relative to No Adapt. Severities 1, 3, and 5 were evaluated.}
\label{fig:severity}
\end{figure}

\noindent\textbf{Effect of corruption severity.}
ELaTTA refines features that still carry class information and cannot recover information the backbone has already lost. \Fref{fig:severity} shows this on the Contrast corruption at severities 1, 3, and 5. At severities 1 and 3, ELaTTA improves ResNet-50 by 7.44 and 3.08 points and MobileNet-V4 by 10.24 and 8.69 points. At severity 5, unadapted accuracy is 0.72\% and 3.84\%, the class signal is essentially gone, and ELaTTA changes accuracy by $-0.47$ and $-1.06$ points. The gain thus shrinks as severity increases and turns slightly negative at severity 5. Other corruptions on which the unadapted model is close to zero, such as Glass Blur on MobileNet-V4 (\Tref{tab:diverse_networks}), show no gain either.

\section{Quantization of ELaTTA}
\label{sec:x_quant}
ELaTTA concentrates online adaptation in a low-dimensional coefficient vector and its search procedure. We study the sensitivity of these components to numerical precision through two quantized variants (QELaTTA). Both apply latent corrections within the fixed source-derived subspace while keeping the model weights fixed.

\noindent\textbf{Quantized variants.}
Let $\mathbf{p}\in\mathbb{R}^k$ denote the coefficient vector used for latent correction. The two variants are
\begin{definition}
  QELaTTA-V1. Before scoring each candidate, we quantize $\mathbf{p}$ into a \emph{1-bit} representation, so each coefficient can take only two values. The candidate entropy is then evaluated using the quantized coefficients.
\end{definition}
\begin{definition}
    QELaTTA-V2. Building on QELaTTA-V1, we further quantize the whole CMA-ES procedure by using \emph{fixed-point} arithmetic with configurable bit-widths, aiming to reduce reliance on high-precision floating-point support and preserve the effectiveness of the optimization dynamics.
\end{definition}

\begin{table*}[t]
\caption{Accuracy (\%) of ELaTTA and its quantized variants on ImageNet-C at severity 5 with ViT-Base under strict single-instance adaptation. QELaTTA-V2 ($x$b$y$) uses $x$-bit fixed-point arithmetic with $y$ integer bits for CMA-ES.}
\label{tab:quan_1}
\centering
\resizebox{\textwidth}{!}{
\begin{tabular}{l ccc cccc cccc cccc >{\columncolor{gray!20}}c}
\toprule
\multirow{2}{*}{Method} & \multicolumn{3}{c}{Noise} & \multicolumn{4}{c}{Blur} & \multicolumn{4}{c}{Weather} & \multicolumn{4}{c}{Digital} & Average\\
& Gauss. & Shot & Impl. & Defoc. & Glass & Motion & Zoom & Snow & Frost & Fog & Brit. & Contr. & Elas. & Pix. & JPEG & Acc. \\
\midrule
ELaTTA     & 58.77 & 59.66 & 59.50 & 49.30 & 36.08 & 55.35 & 46.34 & 65.21 & 66.40 & 67.66 & 80.21 & 35.96 & 47.61 & 69.55 & 69.68 & 57.82 \\
\midrule
QELaTTA-V1 & 59.41 & 60.15 & 60.09 & 49.86 & 36.48 & 55.94 & 46.70 & 65.60 & 66.74 & 60.90 & 80.46 & 34.55 & 47.72 & 69.83 & 70.01 & 57.63 \\
\midrule
QELaTTA-V2 (8b5) & 56.16 & 57.11 & 56.93 & 47.17 & 35.14 & 53.60 & 44.76 & 63.18 & 63.94 & 67.60 & 78.37 & 33.03 & 46.26 & 67.48 & 67.49 & 55.88 \\
QELaTTA-V2 (8b4) & 56.75 & 57.75 & 57.35 & 47.63 & 35.32 & 53.99 & 45.12 & 63.60 & 64.39 & 68.34 & 78.66 & 33.71 & 46.51 & 67.83 & 67.88 & 56.32 \\
QELaTTA-V2 (8b3) & 56.73 & 57.61 & 57.42 & 47.63 & 35.35 & 53.90 & 45.04 & 63.49 & 64.35 & 68.07 & 78.65 & 33.16 & 46.53 & 67.75 & 67.77 & 56.23 \\
QELaTTA-V2 (8b2) & 56.64 & 57.59 & 57.37 & 47.55 & 35.37 & 53.99 & 45.08 & 63.52 & 64.28 & 67.76 & 78.56 & 33.41 & 46.49 & 67.72 & 67.81 & 56.21 \\
QELaTTA-V2 (4b4) & 55.06 & 55.88 & 55.62 & 45.53 & 34.41 & 52.50 & 43.96 & 62.39 & 61.66 & 64.82 & 77.59 & 33.29 & 45.46 & 66.48 & 66.54 & 54.75 \\
QELaTTA-V2 (4b2) & 55.41 & 56.35 & 56.15 & 46.56 & 34.78 & 52.93 & 44.24 & 62.44 & 62.85 & 65.84 & 77.82 & 32.10 & 45.78 & 66.76 & 66.72 & 55.12 \\
\bottomrule
\end{tabular}
}
\end{table*}

\begin{table}[t]
\caption{Accuracy (\%) of ELaTTA and its quantized variants on ImageNet-V2, ImageNet-R, and ImageNet-Sketch with ViT-Base under strict single-instance adaptation. QELaTTA-V2 ($x$b$y$) uses $x$-bit fixed-point arithmetic with $y$ integer bits for CMA-ES.}
\label{tab:quan_2}
\centering
\begin{tabular}{l c c c >{\columncolor{gray!20}}c}
\toprule
\multirow{2}{*}{Method} & \multicolumn{4}{c}{Accuracy (\%)} \\
& V2 & R & Sketch & Avg. \\
\midrule
ELaTTA           & 78.15 & 65.29 & 47.73 & 63.72 \\
\midrule
QELaTTA-V1       & 77.46 & 63.31 & 46.79 & 62.52 \\
\midrule
QELaTTA-V2 (8b5) & 76.94 & 62.02 & 46.33 & 61.76 \\
QELaTTA-V2 (8b4) & 77.21 & 62.06 & 46.37 & 61.88 \\
QELaTTA-V2 (8b3) & 76.86 & 61.45 & 45.99 & 61.43 \\
QELaTTA-V2 (8b2) & 76.26 & 60.42 & 45.39 & 60.69 \\
QELaTTA-V2 (4b4) & 75.17 & 57.70 & 44.65 & 59.17 \\
QELaTTA-V2 (4b2) & 75.46 & 59.42 & 44.88 & 59.92 \\
\bottomrule
\end{tabular}
\end{table}

\noindent\textbf{Experimental results.}
\Tref{tab:quan_1} and \Tref{tab:quan_2} report accuracy on four visual benchmarks. QELaTTA-V1 reaches 57.63\% on ImageNet-C and 62.52\% averaged over ImageNet-V2, ImageNet-R, and ImageNet-Sketch, decreases of 0.19 and 1.20 percentage points from ELaTTA. Binary coefficients thus retain most of the average adaptation gain. Their effect varies across shifts: accuracy improves over ELaTTA on 13 of the 15 ImageNet-C corruptions, but drops by 6.76 points on Fog and 1.41 points on Contrast.

Quantizing the CMA-ES procedure introduces a larger accuracy loss. Among the tested QELaTTA-V2 settings, 8b4 gives the highest averages: 56.32\% on ImageNet-C and 61.88\% over ImageNet-V2/R/Sketch. These remain above the corresponding No Adapt averages of 55.03\% and 59.96\%. The 4-bit settings reduce accuracy further; on ImageNet-C, 4b4 reaches 54.75\%, below No Adapt, while 4b2 reaches 55.12\%. The differences between settings at the same total bit-width also show that the allocation of integer bits affects search performance. Together, these results characterize the accuracy trade-off of binary coefficient representation and fixed-point search, providing a basis for choosing numerical precision in future hardware implementations.

\section{Conclusion and Limitation}
We presented ELaTTA, a forward-only framework for single-instance TTA that searches low-dimensional coefficients in a source-derived principal latent subspace while keeping all model weights fixed. The source basis defines the correction directions, and covariance-adaptive Gaussian search uses entropy feedback to select a correction within a finite evaluation budget. Experiments on five visual benchmarks show strong adaptation performance with low computation and memory costs, supported by component ablations, source-cache stability analysis, and improved calibration on ImageNet-C. Keyword spotting demonstrates applicability beyond vision, and ViT-Base deployment on a Raspberry Pi 5 confirms practical on-device feasibility. Quantization experiments show that binary coefficients retain much of the adaptation gain, while lower-precision search introduces larger accuracy losses. The method's effectiveness depends on the chosen subspace, search budget, and class information retained by the encoder. Severe corruption and confident errors remain challenges for entropy-based latent correction. Future work will explore adaptive search budgets and hardware-aware implementations to further improve the accuracy--efficiency trade-off of on-device adaptation.

\backmatter

\section*{Statements and Declarations}
\textbf{Competing interests.}
The authors have no conflict of interest to declare that are relevant to the content of this article.

\noindent\textbf{Data availability.}
All data supporting the findings of this study are publicly available and cited in \Sref{sec:exp_setup}.

\noindent\textbf{Authors' contributions.}
Xinyu Luo led the algorithm design, conducted the experiments, and drafted the manuscript. Jie Liu contributed theoretical insights to the algorithm design. Kecheng Chen contributed to the algorithm design, manuscript writing, and revision. Yihua Shao designed the software framework for experimental evaluation. Junyi Yang designed the edge-device experiments. Bo Ding revised the manuscript. Nicu Sebe, Arindam Basu and Haoliang Li guided algorithm design, provided computational resources, and revised the manuscript.

\bibliography{sn-bibliography}

@String(IJCV = {Int. J. Comput. Vis.})

@String(ICCV= {Int. Conf. Comput. Vis.})

@String(IJCV  = {IJCV})

@String(ICCV  = {ICCV})

@inproceedings{wang2021tent,
  title={Tent: Fully Test-Time Adaptation by Entropy Minimization},
  author={Wang, Dequan and Shelhamer, Evan and Liu, Shaoteng and Olshausen, Bruno and Darrell, Trevor},
  booktitle={International Conference on Learning Representations},
  year={2021},
  url={https://openreview.net/forum?id=uXl3bZLkr3c}
}

@article{zhang2022memo,
  title={Memo: Test time robustness via adaptation and augmentation},
  author={Zhang, Marvin and Levine, Sergey and Finn, Chelsea},
  journal={Advances in neural information processing systems},
  volume={35},
  pages={38629--38642},
  year={2022}
}

@article{grandvalet2004semi,
  title={Semi-supervised learning by entropy minimization},
  author={Grandvalet, Yves and Bengio, Yoshua},
  journal={Advances in neural information processing systems},
  volume={17},
  year={2004}
}

@inproceedings{chen2025test,
  title={Test-time Adaptation for Foundation Medical Segmentation Model without Parametric Updates},
  author={Chen, Kecheng and Luo, Xinyu and Qin, Tiexin and Liu, Jie and Liu, Hui and Lee, Victor Ho Fun and Yan, Hong and Li, Haoliang},
  booktitle={Proceedings of the IEEE/CVF International Conference on Computer Vision},
  year={2025}
}

@article{shannon1948mathematical,
  title={A mathematical theory of communication},
  author={Shannon, Claude E},
  journal={The Bell system technical journal},
  volume={27},
  number={3},
  pages={379--423},
  year={1948},
  publisher={Nokia Bell Labs}
}

@article{iwasawa2021test,
  title={Test-time classifier adjustment module for model-agnostic domain generalization},
  author={Iwasawa, Yusuke and Matsuo, Yutaka},
  journal={Advances in Neural Information Processing Systems},
  volume={34},
  pages={2427--2440},
  year={2021}
}

@article{luo2025space,
  title={SPACE: SPike-Aware Consistency Enhancement for Test-Time Adaptation in Spiking Neural Networks},
  author={Luo, Xinyu and Chen, Kecheng and Sun, Pao-Sheng Vincent and Tian, Chris Xing and Basu, Arindam and Li, Haoliang},
  journal={Advances in Neural Information Processing Systems},
  year={2025}
}

@article{hansen2016cma,
  title={The CMA evolution strategy: A tutorial},
  author={Hansen, Nikolaus},
  journal={arXiv preprint arXiv:1604.00772},
  year={2016}
}

@inproceedings{li2024flexnn,
  title={Flexnn: Efficient and adaptive dnn inference on memory-constrained edge devices},
  author={Li, Xiangyu and Li, Yuanchun and Li, Yuanzhe and Cao, Ting and Liu, Yunxin},
  booktitle={Proceedings of the 30th Annual International Conference on Mobile Computing and Networking},
  pages={709--723},
  year={2024}
}

@article{hendrycks2019robustness,
  title={Benchmarking Neural Network Robustness to Common Corruptions and Perturbations},
  author={Dan Hendrycks and Thomas Dietterich},
  journal={Proceedings of the International Conference on Learning Representations},
  year={2019}
}

@article{hendrycks2021many,
  title={The Many Faces of Robustness: A Critical Analysis of Out-of-Distribution Generalization},
  author={Dan Hendrycks and Steven Basart and Norman Mu and Saurav Kadavath and Frank Wang and Evan Dorundo and Rahul Desai and Tyler Zhu and Samyak Parajuli and Mike Guo and Dawn Song and Jacob Steinhardt and Justin Gilmer},
  journal={ICCV},
  year={2021}
}

@inproceedings{recht2019imagenet,
  title={Do imagenet classifiers generalize to imagenet?},
  author={Recht, Benjamin and Roelofs, Rebecca and Schmidt, Ludwig and Shankar, Vaishaal},
  booktitle={International conference on machine learning},
  pages={5389--5400},
  year={2019},
  organization={PMLR}
}

@inproceedings{sun2020test,
  title={Test-time training with self-supervision for generalization under distribution shifts},
  author={Sun, Yu and Wang, Xiaolong and Liu, Zhuang and Miller, John and Efros, Alexei and Hardt, Moritz},
  booktitle={International conference on machine learning},
  pages={9229--9248},
  year={2020},
  organization={PMLR}
}

@inproceedings{wang2022continual,
  title={Continual test-time domain adaptation},
  author={Wang, Qin and Fink, Olga and Van Gool, Luc and Dai, Dengxin},
  booktitle={Proceedings of the IEEE/CVF Conference on Computer Vision and Pattern Recognition},
  pages={7201--7211},
  year={2022}
}

@inproceedings{niu2022efficient,
  title={Efficient test-time model adaptation without forgetting},
  author={Niu, Shuaicheng and Wu, Jiaxiang and Zhang, Yifan and Chen, Yaofo and Zheng, Shijian and Zhao, Peilin and Tan, Mingkui},
  booktitle={International conference on machine learning},
  pages={16888--16905},
  year={2022},
  organization={PMLR}
}

@inproceedings{lim2023ttn,
  title={TTN: A domainshift aware batch normalization in test-time adaptation},
  author={Lim, Hyesu and Kim, Byeonggeun and Choo, Jaegul and Choi, Sungha},
  booktitle={11th International Conference on Learning Representations},
  year={2023}
}

@inproceedings{boudiaf2022parameter,
  title={Parameter-free online test-time adaptation},
  author={Boudiaf, Malik and Mueller, Romain and Ben Ayed, Ismail and Bertinetto, Luca},
  booktitle={Proceedings of the IEEE/CVF Conference on Computer Vision and Pattern Recognition},
  pages={8344--8353},
  year={2022}
}

@article{schneider2020improving,
  title={Improving robustness against common corruptions by covariate shift adaptation},
  author={Schneider, Steffen and Rusak, Evgenia and Eck, Luisa and Bringmann, Oliver and Brendel, Wieland and Bethge, Matthias},
  journal={Advances in neural information processing systems},
  volume={33},
  pages={11539--11551},
  year={2020}
}

@inproceedings{niu2024test,
  title={Test-Time Model Adaptation with Only Forward Passes},
  author={Niu, Shuaicheng and Miao, Chunyan and Chen, Guohao and Wu, Pengcheng and Zhao, Peilin},
  booktitle = {The International Conference on Machine Learning},
  year = {2024}
}

@article{khurana2021sita,
  title={Sita: Single image test-time adaptation},
  author={Khurana, Ansh and Paul, Sujoy and Rai, Piyush and Biswas, Soma and Aggarwal, Gaurav},
  journal={arXiv preprint arXiv:2112.02355},
  year={2021}
}

@article{gong2022note,
  title={Note: Robust continual test-time adaptation against temporal correlation},
  author={Gong, Taesik and Jeong, Jongheon and Kim, Taewon and Kim, Yewon and Shin, Jinwoo and Lee, Sung-Ju},
  journal={Advances in Neural Information Processing Systems},
  volume={35},
  pages={27253--27266},
  year={2022}
}

@article{hu2021mixnorm,
  title={Mixnorm: Test-time adaptation through online normalization estimation},
  author={Hu, Xuefeng and Uzunbas, Gokhan and Chen, Sirius and Wang, Rui and Shah, Ashish and Nevatia, Ram and Lim, Ser-Nam},
  journal={arXiv preprint arXiv:2110.11478},
  year={2021}
}

@inproceedings{piczak2015esc,
  title={ESC: Dataset for environmental sound classification},
  author={Piczak, Karol J},
  booktitle={Proceedings of the 23rd ACM international conference on Multimedia},
  pages={1015--1018},
  year={2015}
}

@article{warden2018speech,
  title={Speech commands: A dataset for limited-vocabulary speech recognition},
  author={Warden, Pete},
  journal={arXiv preprint arXiv:1804.03209},
  year={2018}
}

@article{wang2019learning,
  title={Learning robust global representations by penalizing local predictive power},
  author={Wang, Haohan and Ge, Songwei and Lipton, Zachary and Xing, Eric P},
  journal={Advances in neural information processing systems},
  volume={32},
  year={2019}
}

@inproceedings{dosovitskiyimage,
  title={An Image is Worth 16x16 Words: Transformers for Image Recognition at Scale},
  author={Dosovitskiy, Alexey and Beyer, Lucas and Kolesnikov, Alexander and Weissenborn, Dirk and Zhai, Xiaohua and Unterthiner, Thomas and Dehghani, Mostafa and Minderer, Matthias and Heigold, Georg and Gelly, Sylvain and others},
  booktitle={International Conference on Learning Representations},
  year={2021},
}

@inproceedings{he2016deep,
  title={Deep residual learning for image recognition},
  author={He, Kaiming and Zhang, Xiangyu and Ren, Shaoqing and Sun, Jian},
  booktitle={Proceedings of the IEEE conference on computer vision and pattern recognition},
  pages={770--778},
  year={2016}
}

@article{ILSVRC15,
Author = {Olga Russakovsky and Jia Deng and Hao Su and Jonathan Krause and Sanjeev Satheesh and Sean Ma and Zhiheng Huang and Andrej Karpathy and Aditya Khosla and Michael Bernstein and Alexander C. Berg and Li Fei-Fei},
Title = {{ImageNet Large Scale Visual Recognition Challenge}},
Year = {2015},
journal   = {International Journal of Computer Vision (IJCV)},
doi = {10.1007/s11263-015-0816-y},
volume={115},
number={3},
pages={211-252}
}

@inproceedings{niutowards,
  title={Towards Stable Test-time Adaptation in Dynamic Wild World},
  author={Niu, Shuaicheng and Wu, Jiaxiang and Zhang, Yifan and Wen, Zhiquan and Chen, Yaofo and Zhao, Peilin and Tan, Mingkui},
  booktitle={The Eleventh International Conference on Learning Representations},
  year={2023}
}

@inproceedings{tan2019efficientnet,
  title={Efficientnet: Rethinking model scaling for convolutional neural networks},
  author={Tan, Mingxing and Le, Quoc},
  booktitle={International conference on machine learning},
  pages={6105--6114},
  year={2019},
  organization={PMLR}
}

@inproceedings{qin2024mobilenetv4,
  title={MobileNetV4: universal models for the mobile ecosystem},
  author={Qin, Danfeng and Leichner, Chas and Delakis, Manolis and Fornoni, Marco and Luo, Shixin and Yang, Fan and Wang, Weijun and Banbury, Colby and Ye, Chengxi and Akin, Berkin and others},
  booktitle={European Conference on Computer Vision},
  pages={78--96},
  year={2024},
  organization={Springer}
}

@inproceedings{darestani2022test,
  title={Test-time training can close the natural distribution shift performance gap in deep learning based compressed sensing},
  author={Darestani, Mohammad Zalbagi and Liu, Jiayu and Heckel, Reinhard},
  booktitle={International conference on machine learning},
  pages={4754--4776},
  year={2022},
  organization={PMLR}
}

@article{liang2025comprehensive,
  title={A comprehensive survey on test-time adaptation under distribution shifts},
  author={Liang, Jian and He, Ran and Tan, Tieniu},
  journal={International Journal of Computer Vision},
  volume={133},
  number={1},
  pages={31--64},
  year={2025},
  publisher={Springer}
}

@inproceedings{djelouah2019content,
  title={Content adaptive optimization for neural image compression},
  author={Djelouah, JCMSA and Schroers, Christopher},
  booktitle={Proc. IEEE Comput. Soc. Conf. Comput. Vis. Pattern Recognit},
  volume={2},
  pages={1--5},
  year={2019}
}

@inproceedings{shen2023dec,
  title={Dec-adapter: Exploring efficient decoder-side adapter for bridging screen content and natural image compression},
  author={Shen, Sheng and Yue, Huanjing and Yang, Jingyu},
  booktitle={Proceedings of the IEEE/CVF International Conference on Computer Vision},
  pages={12887--12896},
  year={2023}
}

@inproceedings{shen2020interpreting,
  title={Interpreting the latent space of gans for semantic face editing},
  author={Shen, Yujun and Gu, Jinjin and Tang, Xiaoou and Zhou, Bolei},
  booktitle={Proceedings of the IEEE/CVF conference on computer vision and pattern recognition},
  pages={9243--9252},
  year={2020}
}

@article{vahdat2021score,
  title={Score-based generative modeling in latent space},
  author={Vahdat, Arash and Kreis, Karsten and Kautz, Jan},
  journal={Advances in neural information processing systems},
  volume={34},
  pages={11287--11302},
  year={2021}
}

@inproceedings{huselective,
  title={Selective Label Enhancement Learning for Test-Time Adaptation},
  author={Hu, Yihao and Qiao, Congyu and Geng, Xin and Xu, Ning},
  booktitle={The Thirteenth International Conference on Learning Representations},
  year={2025}
}

@inproceedings{guo2017calibration,
  title={On calibration of modern neural networks},
  author={Guo, Chuan and Pleiss, Geoff and Sun, Yu and Weinberger, Kilian Q},
  booktitle={International conference on machine learning},
  pages={1321--1330},
  year={2017},
  organization={PMLR}
}

@inproceedings{roid,
  title={Universal test-time adaptation through weight ensembling, diversity weighting, and prior correction},
  author={Marsden, Robert A and D{\"o}bler, Mario and Yang, Bin},
  booktitle={Proceedings of the IEEE/CVF Winter Conference on Applications of Computer Vision},
  pages={2555--2565},
  year={2024}
}

@inproceedings{zoa,
  title={Test-time model adaptation for quantized neural networks},
  author={Deng, Zeshuai and Chen, Guohao and Niu, Shuaicheng and Luo, Hui and Zhang, Shuhai and Yang, Yifan and Chen, Renjie and Luo, Wei and Tan, Mingkui},
  booktitle={Proceedings of the 33rd ACM International Conference on Multimedia},
  pages={7258--7267},
  year={2025}
}

@inproceedings{yang2024versatile,
  title={A versatile framework for continual test-time domain adaptation: Balancing discriminability and generalizability},
  author={Yang, Xu and Chen, Xuan and Li, Moqi and Wei, Kun and Deng, Cheng},
  booktitle={Proceedings of the IEEE/CVF Conference on Computer Vision and Pattern Recognition},
  pages={23731--23740},
  year={2024}
}

@inproceedings{peng2019moment,
  title={Moment matching for multi-source domain adaptation},
  author={Peng, Xingchao and Bai, Qinxun and Xia, Xide and Huang, Zijun and Saenko, Kate and Wang, Bo},
  booktitle={Proceedings of the IEEE/CVF international conference on computer vision},
  pages={1406--1415},
  year={2019}
}

@inproceedings{lee2024becotta,
  title={BECoTTA: Input-dependent Online Blending of Experts for Continual Test-time Adaptation},
  author={Lee, Daeun and Yoon, Jaehong and Hwang, Sung Ju},
  booktitle={International Conference on Machine Learning},
  pages={27072--27093},
  year={2024},
  organization={PMLR}
}

@inproceedings{ma2025surgeon,
  title={SURGEON: Memory-Adaptive Fully Test-Time Adaptation via Dynamic Activation Sparsity},
  author={Ma, Ke and Tang, Jiaqi and Guo, Bin and Dang, Fan and Liu, Sicong and Zhu, Zhui and Wu, Lei and Fang, Cheng and Chen, Ying-Cong and Yu, Zhiwen and others},
  booktitle={Proceedings of the Computer Vision and Pattern Recognition Conference},
  pages={30514--30523},
  year={2025}
}

@inproceedings{song2023ecotta,
  title={Ecotta: Memory-efficient continual test-time adaptation via self-distilled regularization},
  author={Song, Junha and Lee, Jungsoo and Kweon, In So and Choi, Sungha},
  booktitle={Proceedings of the IEEE/CVF Conference on Computer Vision and Pattern Recognition},
  pages={11920--11929},
  year={2023}
}

@inproceedings{hong2023mecta,
  title={Mecta: Memory-economic continual test-time model adaptation},
  author={Hong, Junyuan and Lyu, Lingjuan and Zhou, Jiayu and Spranger, Michael},
  booktitle={2023 International Conference on Learning Representations},
  year={2023}
}

@inproceedings{mirza2022norm,
  title={The norm must go on: Dynamic unsupervised domain adaptation by normalization},
  author={Mirza, M Jehanzeb and Micorek, Jakub and Possegger, Horst and Bischof, Horst},
  booktitle={Proceedings of the IEEE/CVF conference on computer vision and pattern recognition},
  pages={14765--14775},
  year={2022}
}

@article{yang202533,
  title={A 33.6--136.2-TOPS/W Nonlinear Analog Computing-in-Memory Macro for Multi-Bit LSTM Accelerator in 65-nm CMOS},
  author={Yang, Junyi and Luo, Xinyu and Ke, Ye and Wang, Zheng and Shang, Hongyang and Dong, Shuai and Fu, Zhengnan and Yang, Xiaofeng and Liu, Hongjie and Basu, Arindam},
  journal={IEEE Journal of Solid-State Circuits},
  year={2025},
  publisher={IEEE}
}

@book{golub2013matrix,
  title={Matrix computations},
  author={Golub, Gene H and Van Loan, Charles F},
  year={2013},
  publisher={JHU press}
}

@book{horn2012matrix,
  title={Matrix analysis},
  author={Horn, Roger A and Johnson, Charles R},
  year={2012},
  publisher={Cambridge university press}
}

@article{zhang2024unsupervised,
  title={Unsupervised test-time adaptation learning for effective hyperspectral image super-resolution with unknown degeneration},
  author={Zhang, Lei and Nie, Jiangtao and Wei, Wei and Zhang, Yanning},
  journal={IEEE Transactions on Pattern Analysis and Machine Intelligence},
  volume={46},
  number={7},
  pages={5008--5025},
  year={2024},
  publisher={IEEE}
}

@article{su2024revisiting,
  title={Revisiting realistic test-time training: Sequential inference and adaptation by anchored clustering regularized self-training},
  author={Su, Yongyi and Xu, Xun and Li, Tianrui and Jia, Kui},
  journal={IEEE Transactions on Pattern Analysis and Machine Intelligence},
  volume={46},
  number={8},
  pages={5524--5540},
  year={2024},
  publisher={IEEE}
}

@article{tan2025uncertainty,
  title={Uncertainty-calibrated test-time model adaptation without forgetting},
  author={Tan, Mingkui and Chen, Guohao and Wu, Jiaxiang and Zhang, Yifan and Chen, Yaofo and Zhao, Peilin and Niu, Shuaicheng},
  journal={IEEE Transactions on Pattern Analysis and Machine Intelligence},
  year={2025},
  publisher={IEEE}
}

@article{wang2025search,
  title={In search of lost online test-time adaptation: A survey},
  author={Wang, Zixin and Luo, Yadan and Zheng, Liang and Chen, Zhuoxiao and Wang, Sen and Huang, Zi},
  journal={International Journal of Computer Vision},
  volume={133},
  number={3},
  pages={1106--1139},
  year={2025},
  publisher={Springer}
}

@inproceedings{lee2024entropy,
  title={Entropy is not enough for test-time adaptation: From the perspective of disentangled factors},
  author={Lee, Jonghyun and Jung, Dahuin and Lee, Saehyung and Park, Junsung and Shin, Juhyeon and Hwang, Uiwon and Yoon, Sungroh},
  booktitle={The Twelfth International Conference on Learning Representations},
  year={2024}
}

@inproceedings{zhang2026curvature,
  title={Curvature-Aware Zeroth-Order Optimization for Memory-Efficient Test-Time Adaptation},
  author={Zhang, Junming and Yin, Shuyu and Liu, Peilin and Ying, Rendong and Wen, Fei},
  booktitle={Proceedings of the IEEE/CVF Conference on Computer Vision and Pattern Recognition},
  pages={836--846},
  year={2026}
}

@article{jia2024tinytta,
  title={Tinytta: Efficient test-time adaptation via early-exit ensembles on edge devices},
  author={Jia, Hong and Kwon, Young D and Orsino, Alessio and Dang, Ting and Talia, Domenico and Mascolo, Cecilia},
  journal={Advances in Neural Information Processing Systems},
  volume={37},
  pages={43274--43299},
  year={2024}
}

\end{document}